\documentclass[runningheads]{llncs}
\usepackage{graphicx}
\usepackage{amsmath,amssymb} 
\usepackage{subcaption}
\usepackage{makecell}

\usepackage{capt-of}
\usepackage{multirow}
\usepackage{arydshln}
\usepackage{lipsum}
\newcommand\blfootnote[1]{%
  \begingroup
  \renewcommand\thefootnote{}\footnote{#1}%
  \addtocounter{footnote}{-1}%
  \endgroup
}

\begin{document}
\title{Multi-Scale Convolutions for Learning Context Aware Feature Representations}
\titlerunning{Multi-Scale Convolutions for Context Aware Feature Representations}
%
\author{Nikolai Ufer\inst{*} \and
Kam To Lui\inst{*} \and
Katja Schwarz \and
Paul Warkentin \and \\
Bj\"orn Ommer
}
\authorrunning{N. Ufer, K. T. Lui, K. Schwarz, P. Warkentin, B. Ommer}
%
\institute{
Heidelberg University, HCI / IWR, Germany \\
\email{
nikolai.ufer@iwr.uni-heidelberg.de \\ 
kamto.lui@alumni.uni-heidelberg.de \\
ommer@uni-heidelberg.de}
}
\maketitle              

	\begin{abstract}
	Finding semantic correspondences is a challenging problem. 
    With the breakthrough of CNNs stronger features are available for tasks like classification but not specifically for the requirements of semantic matching.
    In the following we present a weakly supervised metric learning approach which generates stronger features by encoding far more context than previous methods.
    First, we generate more suitable training data using a geometrically informed correspondence mining method which is less prone to spurious matches and requires only image category labels as supervision.
    Second, we introduce a new convolutional layer which is a learned mixture of differently strided convolutions and allows the network 
    to encode implicitly more context while preserving matching accuracy at the same time.
    The strong geometric encoding on the feature side enables us to learn a semantic flow network, which generates more natural deformations than parametric transformation based models and is able to jointly predict foreground regions at the same time.
    Our semantic flow network
    outperforms current state-of-the-art on several semantic matching benchmarks and the learned features show astonishing performance regarding simple nearest neighbor matching.
	\end{abstract}

	\section{Introduction}
	\blfootnote{* Both authors contributed equally.}
Estimating correspondences between images is one of the main problems in computer vision with application in optical flow, stereo matching, structure from-motion, SLAM and segmentation. 
Early approaches have focused on finding correspondence between images depicting identical objects or scenes from different viewpoints. 
With increasingly better feature representations, recent work focuses on estimating correspondences between different instances of the same semantic category \cite{Liu2011SiftApplications,Kim2013DeformableCorrespondences}, 
which is much more challenging due to the severe visual intra-class variations.  
In the literature this problem is also denoted as semantic matching or semantic flow estimation for the task of finding dense correspondences.

In general, matching approaches consist of two main parts: 
First, a feature extraction method, usually a convolutional network, and second, a geometric model, which estimates a global transformation based on local feature similarities.
Current state-of-the-art methods try to learn both parts jointly but suffer from three limitations:
1) The feature representation is often based on standard image classification networks, which are not well suited for finding local correspondences. One important reason is that these networks have only a small effective receptive field \cite{Luo2017UnderstandingNetworks} and cannot resolve matching ambiguities using larger context.
2) The majority of approaches \cite{Rocco2017End-to-endAlignment,Rocco2017ConvolutionalMatching,Han2017SCNet:Correspondence} utilize rigid alignment models, like a TPS \cite{bookstein1989principal}, which aggregate local feature similarities and provide a strong spatial regularization.
However, these models are not able to cope with complex non-linear deformations and provide only rough alignments.
In contrast non-parametric transformation models try to estimate pixel-wise accurate flow fields
\cite{ZhouTinghuiandKrahenbuhlPhilippandAubryMathieuandHuangQixingandEfros2016LearningConsistency,kim2018rtns}.
But learning these models requires a large amount of data due to their many degrees of freedom.
3)  
A lot of progress has been made in the area of image co-segmentation where the foreground objects in two or more images are predicted. 
Recent semantic matching algorithms neglect this aspect: they find pixel-wise correspondences but do not predict which pixels in the two images belongs to the foreground object.

In order to address these limitations we introduce the following innovations.
First, we introduce a geometrically informed correspondence mining procedure, which generates correspondences on a patch level which 
can often be resolved only on a higher contextual scale. 
%
\newcommand{\graphwidth}{.98\linewidth}
\newcommand{\graphheight}{1.35cm}
\newcommand{\skipwidth}{.05cm}
\begin{figure}[!t]
\centering
\begin{subfigure}[b]{.14\linewidth}  
    \centering
    \includegraphics[width=.985\linewidth,height=\graphheight]{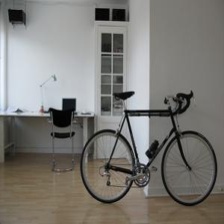} \\
    \vspace{0.05cm}
    \includegraphics[width=\graphwidth,height=\graphheight]{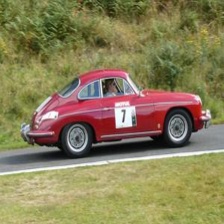} \\
    \vspace{0.05cm}
    \includegraphics[width=.985\linewidth,height=\graphheight]{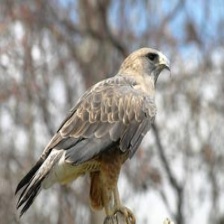}
	\vspace{-0.5cm}
	\caption*{\small Source}
\end{subfigure}
\begin{subfigure}[b]{.14\linewidth}  
    \centering
    \includegraphics[width=.985\linewidth,height=\graphheight]{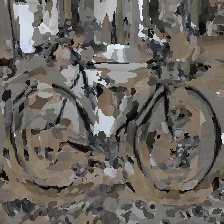} \\
    \vspace{0.05cm}
    \includegraphics[width=\graphwidth,height=\graphheight]{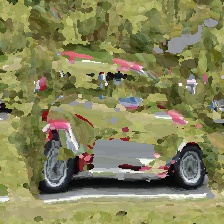} \\
    \vspace{0.05cm}
    \includegraphics[width=.985\linewidth,height=\graphheight]{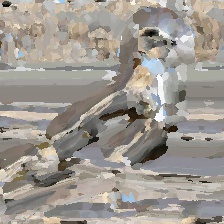}
	\vspace{-0.5cm}
	\caption*{\small SIFT}
\end{subfigure}
\begin{subfigure}[b]{.14\linewidth}  
    \centering
    \includegraphics[width=.985\linewidth,height=\graphheight]{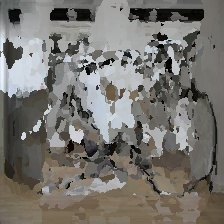} \\
    \vspace{0.05cm}
    \includegraphics[width=\graphwidth,height=\graphheight]{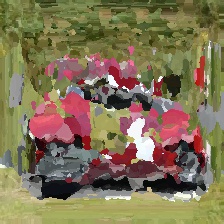} \\
    \vspace{0.05cm}
    \includegraphics[width=.985\linewidth,height=\graphheight]{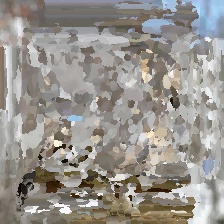}
	\vspace{-0.5cm}
	\caption*{\small FCSS}
\end{subfigure}
\begin{subfigure}[b]{.14\linewidth}  
    \centering
    \includegraphics[width=.985\linewidth,height=\graphheight]{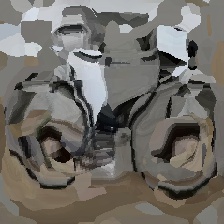} \\
    \vspace{0.05cm}
    \includegraphics[width=\graphwidth,height=\graphheight]{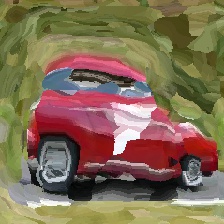} \\
    \vspace{0.05cm}
    \includegraphics[width=.985\linewidth,height=\graphheight]{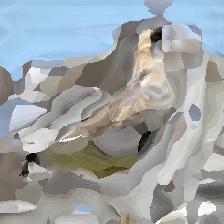}
	\vspace{-0.5cm}
	\caption*{\small WAlign NN}
\end{subfigure}
\begin{subfigure}[b]{.14\linewidth}  
    \centering
    \includegraphics[width=.985\linewidth,height=\graphheight]{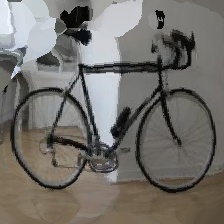} \\
    \vspace{0.05cm}
    \includegraphics[width=\graphwidth,height=\graphheight]{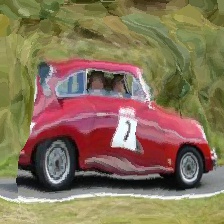} \\
    \vspace{0.05cm}
    \includegraphics[width=.985\linewidth,height=\graphheight]{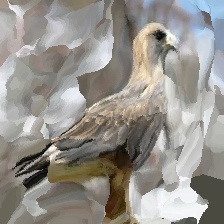} 
	\vspace{-0.5cm}
	\caption*{\small Ours NN}
\end{subfigure}
\begin{subfigure}[b]{.14\linewidth} 
    \centering
    \includegraphics[width=.985\linewidth, height=\graphheight]{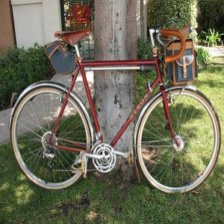}
    \\
    \vspace{0.05cm}
    \includegraphics[width=\graphwidth,height=\graphheight]{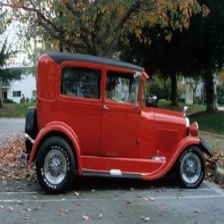} \\
    \vspace{0.05cm}
    \includegraphics[width=.985\linewidth,height=\graphheight]{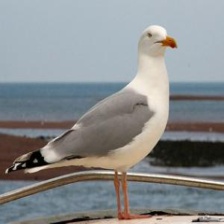}
    \vspace{-0.5cm}
    \caption*{\small Target}
\end{subfigure}
\vspace{-0.10cm}
\caption{
\small
Nearest neighbor matching visualization. 
Previous learned features for semantic matching encode too little context information.
While they allow to find correspondences, they ignore larger context like the shape of the object, see the results of SIFT \cite{loweSift}, FCSS \cite{kim2018fcss} and ResNet-101 features, which are fine-tuned during the weak alignment training \cite{Rocco2017ConvolutionalMatching} (WAlign NN).
In contrast our features are able to encode larger context and resolve the matching ambiguity, which can be seen in the clear outline of the transformed foreground object, see the fifth column.  
} 
\label{fig:first_teaser}
\end{figure}
Second, we introduce a new network layer which consists of a learnable mixture of convolutions with different dilation strides. This new layer allows the network to propagate information from different spatial resolutions through the network in order to increase the size of the effective receptive field and preserve the spatial localization at the same time.
Due to the context enriched features we are able to obtain a strong baseline for dense semantic matching without any transformation model, see Fig. \ref{fig:first_teaser}.
Based on these features the loose geometric regularization of local transformation models can be resolved and we are able to train a semantic flow network in a weakly supervised way which provides more natural warps compared to previous parametric models.
Furthermore, we utilize our mined correspondences to learn a segmentation module in our semantic flow network which is capable to provide pixel-wise foreground predictions.

\section{Related work}

\hspace{\parindent}
\textbf{Dense semantic matching.}
The pioneering work in semantic matching was done by Liu et al. with the development of SIFT Flow \cite{Liu2011SiftApplications} and its extension from Kim et al. \cite{Kim2013DeformableCorrespondences}.
They densely sampled SIFT descriptors and formulated a discrete optimization problem for finding an optimal flow field.
Due to the success of CNNs more recent approaches replaced hand engineered descriptors \cite{Liu2011SiftApplications,Dalal2004HistogramsDetection} with learned features.
First there was a strong focus on finding geometric sensitive feature representations for semantic matching \cite{Choy2016UniversalNetwork,Kim2017FCSS:Correspondence}. 
Choey et al. \cite{Choy2016UniversalNetwork} introduced a fully convolutional Siamese architecture with a contrastive loss. 
And Kim et al. \cite{Kim2017FCSS:Correspondence} proposed a convolutional descriptor based on the principle of local self-similarity, where the geometric alignment was based on hand-engineered models.
More recent approaches combined learning descriptors and geometric alignment models \cite{Ham2016ProposalFlow,Rocco2017ConvolutionalMatching}.
Han et al. \cite{Ham2016ProposalFlow} follow Proposal Flow \cite{Kim2017FCSS:Correspondence} and formulate a convolutional neural network architecture based on region proposals.
Using the feature correlation map generated by a Siamese architecture, Roccoco et al. \cite{Rocco2017End-to-endAlignment,Rocco2017ConvolutionalMatching} perform regression on the parameters of a thin plate spline \cite{bookstein1989principal}.
However, these methods are based on parametric transformation models, which provide a strong geometric regularization and poor spatial alignment.
There are only a few exceptions where trainable networks for semantic flow fields are examined \cite{ZhouTinghuiandKrahenbuhlPhilippandAubryMathieuandHuangQixingandEfros2016LearningConsistency}.
Zhou at al. \cite{ZhouTinghuiandKrahenbuhlPhilippandAubryMathieuandHuangQixingandEfros2016LearningConsistency} propose an encoder-decoder network for dense semantic flow, which is inspired by architectures from optical flow, but it requires 3D CAD models to synthesize a large set of pixel-wise accurate ground truth flow.
Just recently Kim et al. \cite{kim2018rtns} introduced a recurrent network for learning a semantic flow field trained on a set of matching images pairs. 
But they are not capable to perform pixel-wise foreground prediction.
%

\textbf{Weakly supervised training.}
The majority of semantic matching frameworks are trained on strong supervision in the form of key point annotations \cite{Choy2016UniversalNetwork,Han2017SCNet:Correspondence}, 3D models \cite{ZhouTinghuiandKrahenbuhlPhilippandAubryMathieuandHuangQixingandEfros2016LearningConsistency}, bounding box annotations \cite{Kim2017FCSS:Correspondence} or image segmentation \cite{jeon2018parn}. There are only few exceptions, like Novotny et al. \cite{Novotny2017AnchorNet:Matching}, which use image-level labels to learn discriminative part filters matching across different object instances.
But the accuracy of these features are rather limited and restricted to specific object categories.
Another popular approach is to utilize synthetically generated image pairs \cite{Rocco2017End-to-endAlignment,Rocco2017ConvolutionalMatching,WarpnetKanazawaJC16}.
For conducting alignment of fine-grained categories WarpNet \cite{WarpnetKanazawaJC16} estimates the parameters of a thin plate spline using a point transformer and synthetically warped images.
A similar approach is done by Roccoco et al. \cite{Rocco2017End-to-endAlignment}, which regresses the geometric model parameters based on a feature correlation map and train on a large synthetically generated set of image pairs.
In the extension \cite{Rocco2017ConvolutionalMatching} they improve their results by fine tuning with an unsupervised soft-inlier count using provided image pairs.
Another approach for generating training data is to estimate correspondences using a simple one-cycle-constraint \cite{Kim2017FCSS:Correspondence,jeon2018parn}. But this is prone to background clutter and therefore also requires strong supervision in terms of bounding boxes or segmentation masks.
In contrast our method does not required labeled image pairs \cite{Rocco2017End-to-endAlignment,jeon2018parn,kim2018rtns}, synthetic training data \cite{Rocco2017End-to-endAlignment,Rocco2017ConvolutionalMatching}, segmentation masks \cite{jeon2018parn} or bounding boxes \cite{Kim2017FCSS:Correspondence}, but only relies on image category labels as source of supervision.
Therefore, it makes it possible to learn powerful feature representations for semantic matching from large-scale datasets like ImageNet.

\textbf{Context aware feature representation.}
To incorporate more contextual information various approaches have been proposed to enlarge the receptive field size.
For semantic segmentation Long et al. \cite{DBLP:LongZD14} proposed a skip architecture to aggregate multiple scales of deep feature hierarchies. 
Yu et al. \cite{yu2015multi} introduced dilated convolutions, also known as atrous convolution, which increases the internal filter's stride.
Dai et al. \cite{Dai2017DeformableNetworks} proposed an extension, where the grid like sampling pattern was resolved and augmented with an additional offset module which learns the offset from the target task. 
For estimating visual correspondences Wang et al. \cite{wang2016autoscaler} propose a multi-scale pyramid approach, where multi-scale features maps are combined in an optimal way using an attention module. 
%
In contrast we learn a mixture of dilated convolutions for each layer which enables the network to propagate information from different scales through the network preserving the overall spatial accuracy.

\textbf{Contributions.}
There are four main contributions.
(1) We propose a correspondence mining procedure which provides image representations that are better suited for semantic matching compared to training on sparse keypoint annotations.
(2) We propose a new convolutional layer, called multi-scale convolution, which produces highly context aware feature representations for the task of semantic matching.
(3) Our semantic flow network is not only capable to predict accurate semantic flow but also to provide foreground segmentations.
(4) We outperform state-of-the-art methods on challenging benchmark datasets.
And simple nearest neighbor matching with our features show competitive performance compared to state-of-the-art matching methods which utilize sophisticated learned transformations models.

\section{Approach}
In the following we present our new multi-scale convolutional network for learning highly context aware feature representations.
To alleviate tedious manual labeling and to obtain enough training data we introduce a weakly supervised mining procedure which requires only class labels. 
On top of this feature representation we train a semantic flow network, 
which 
provides pixel-wise accurate semantic flow and foreground segmentations at the same time.



%
%
%
%
%
%
%
%
%
%
%

\begin{figure}[!t]
\centering
\begin{minipage}{.45\linewidth}
	\includegraphics[width=.85\linewidth, height=0.47\linewidth]{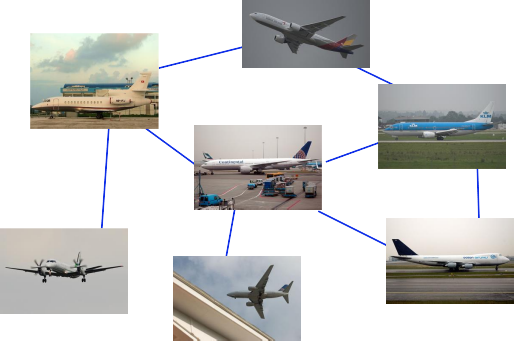}
	\vspace{-0.05cm}
	\caption*{(a)} 
\end{minipage}
\hspace{0.05\linewidth}
\begin{minipage}{.45\linewidth}
	\includegraphics[width=.85\linewidth, height=0.47\linewidth]{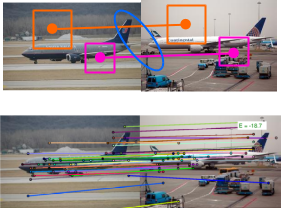}
    \vspace{-0.05cm}
	\caption*{\hspace{-0.8cm}(b)} 
\end{minipage}
\vspace{-0.05cm}
\caption{\small
Overview of our correspondence mining approach. Based on global image features we generate an image graph
to find similar image pairs showing objects from similar type and viewpoint (a).
On these pairs we apply our sparse graph matching on a patch level to find consistent correspondences (b). 
} 
\label{fig:graphmatching}
\end{figure}


%

\subsection{Weakly supervised correspondence mining}
Given a set of images $\mathcal{I}$ from the same category, our weakly supervised mining approach consists of two main steps.
First, using a similarity graph, we identify images depicting semantically similar objects from the same viewpoint and second we utilize a sparse graph matching approach, which finds corresponding regions between a given image pair.

\textbf{Choosing image pairs to match.} 
Mining correspondences between all image pairs is not feasible and not desirable since objects from different viewpoints or semantically different types would lead to spurious matches.
Therefore, we use a simple, yet effective heuristic and construct a k-nearest neighbor graph $\mathcal{G}(\mathcal{I}, \mathcal{N})$, where an edge $(i,j) \in \mathcal{N}$ exists if image $I_j$ is among the k nearest neighbors of image $I_i$. 
We assume that neighboring images are semantically close and depict objects from similar viewpoints.
By selecting all edges we obtain a large set of image pairs on which we apply our graph matching approach for finding corresponding regions.


\textbf{Finding corresponding regions.} 
For finding matching regions we utilize the spatial feature pyramid representation of Girshick et al. \cite{girshick2015deformable} and the sparse graph matching approach of Ufer et al. \cite{Ufer2017DeepMatching}. 
Given an image pair $(I_i, I_j)$ we extract CNN features from the fourth layer of the network for 6 different image resolutions forming a feature pyramid for both images.
Then we extract features in a sliding window manner with a fixed window size, which corresponds to a coverage of both images with patches of different sizes. 
For these patches, which are covering a range from $20 \text{x} 20$ up to $80 \text{x} 80$ pixels in a $224 \text{x} 224$ image, we are searching for corresponding regions.
To reduce the overall number of patches we select a subset in both images using non-maximum suppression on the overall feature activation and a one-cycle consistency constraint.
Based on these selected patches in $I_i$ we search for their $k$ nearest neighbors in the image $I_j$ and obtain matching candidates.
Now, analogously to \cite{Ufer2017DeepMatching,torresani2013dual}, selecting geometric consistent matching candidates in both directions can be formulated as a one-to-one matching problem which leads to a second order Integer Quadratic Program (IQP) which can easily be solved due to the sparsity of the graph with an approximate inference method like TRWS \cite{kolmogorov2006convergent}.
An additional outlier class makes it possible to discard specific matching candidates which are not in agreement with the dominant geometry. 
This results in our final set of patch correspondences which are generated on different scales and contains a large amount of correspondences which can not be resolved by their local appearance.
\begin{figure*}[!t]
\centering
\begin{minipage}{.42\linewidth}
	\includegraphics[width=1.0\linewidth]{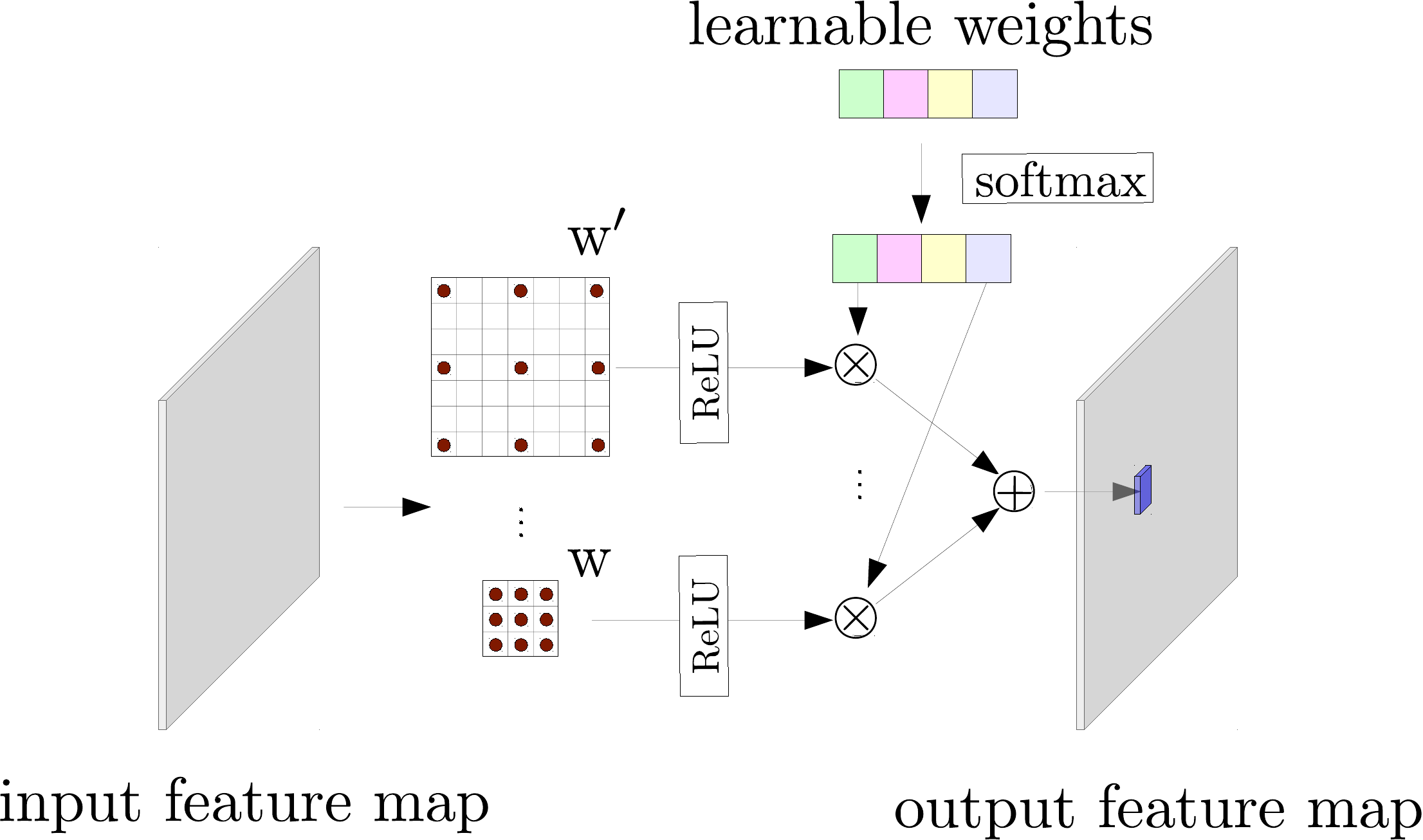}
	\vspace{-0.1cm}
\end{minipage}
\hspace{0.01\linewidth}
\begin{minipage}{.51\linewidth}
	\includegraphics[width=1.0\linewidth]{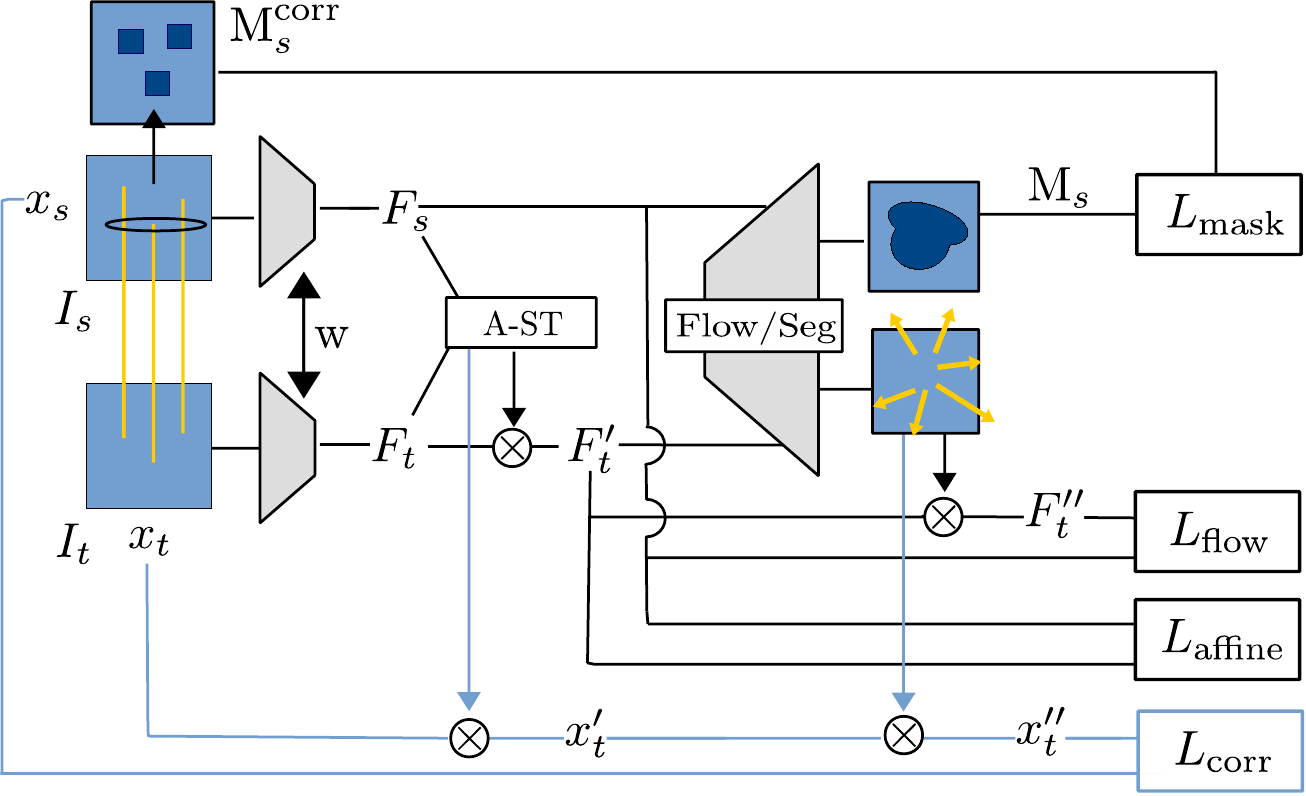}
	\vspace{-0.1cm}
\end{minipage}
\vspace{-0.15cm}
\caption[]{
\small
Schematic illustration of our multi-scale convolution layer (left) and the semantic flow network architecture (right).
} 
\label{fig:network_architecture}
\end{figure*}

\subsection{Feature learning using multi-scale convolutions}
For learning our feature representation we follow the standard metric learning approach with a Siamese network \cite{chopra2005learning}.
Given two images we extract features from intermediate convolutional layers at the centers of corresponding patches and penalize their Euclidean distance using contrastive loss \cite{Choy2016UniversalNetwork,Kim2017FCSS:Correspondence}.
The set of matching pairs is given by our mined correspondences and non-matching pairs are generated by hard negative mining.
We adopt the hard negative mining approach of UCN \cite{Choy2016UniversalNetwork} by searching for nearest neighbors outside a certain radius of corresponding points in the target image. 

%
%
%

\textbf{Multi-scale convolutions.}
For learning context aware feature representations we replace all convolutional layers in the network with a combination of dilated convolutions with different strides, 
which we refer to as multi-scale convolutions.
These dilated convolutions are combined in a weighted sum, where the weights are also learned during back propagation, as illustrated in Fig. \ref{fig:network_architecture}.
To obtain the same dimensions of the output feature maps we adjust the zero padding on the input feature map for each scale separately. 
Each dilated convolution implies a specific receptive field and by optimizing the weighted sum the network learns how to adjust its effective receptive field in each layer for the task of matching.
This is important since the network can use the wider context to resolve ambiguities in semantic matching while a smaller receptive field is still available to maintain localization accuracy.
In the following experiments we will show that adjusting the weights for each layer is crucial since each layer requires a different amount of contextual information.
Our experiments have also shown that applying the non-linear activation before the fusion step provides better results since it prevents the signal of a single scale from suppressing all others.

\subsection{Semantic flow network}
Based on the previously described encoder
we train a semantic flow network which provides dense correspondences and object foreground predictions at the same time.
It consists of three main parts, see Fig. \ref{fig:network_architecture},
which are an affine spatial transformer (A-ST), a flow network and an auxiliary segmentation network.


\textbf{Affine transformer.} 
After extracting the feature maps we fist try to get a rough alignment of the images.
Inspired by the spatial transformer \cite{SpatialTransformerJaderbergSZK15} we concatenate the features $F_s$ and $F_t$ and regress the parameters of an affine transformation from source image $I_s$ to the target image $I_t$ using a localization network. 
It is trained by minimizing the $L2$ loss between features $F_s$ and the transformed target feature $F'_t$, 
\begin{equation}
L_{\text{affine}}=\frac{1}{h_s w_s}\sum^{h_s}_{y=1}\sum^{w_s}_{x=1}{\Vert F_s(x,y)-F'_t(x,y) \Vert_{2}} ,
\end{equation}
where $h_s$ and $w_s$ are the height and width of the source image.
\textbf{Flow network.} 
To get pixel-wise accurate correspondences we utilize a flow network on top.
We follow a similar network architecture as the popular FlowNetSimple model 
\cite{FlowNetFischer}.
The input of the network are the concatenated features $F_s$ and $F'_t$, and the output is 
the flow from source to the affine transformed target image.
The warped target feature $F''_t$ is constructed from $F'_t$ via grid sampling 
and the final loss between features $F_s$ and $F''_t$ is computed by the $L2$ distance
\begin{equation}
L_{\text{flow}}=\frac{1}{h_s w_s}\sum^{h_s}_{y=1}\sum^{w_s}_{x=1}{\Vert F_s(x,y)-F''_t(x,y) \Vert_{2}}  .
\end{equation}
%

\textbf{Correspondence loss.} 
Despite the previously introduced unsupervised feature losses which try to find transformations by mapping similar feature on top of each other, we
utilize the mined correspondences as proxy ground truth and minimize their $L2$ loss,
\begin{equation}
L_{\text{corr}} =  \frac{1}{\sqrt{h_{s}^2 + w_{s}^2}}
\frac{1}{|\Omega|}
\sum_{i \in \Omega} \Vert x_{s,i} - x''_{t,i} \Vert_{2},
\end{equation}
where $(x_{s,i}, x_{t,i})$ are the mined correspondences between the source and target image and the $x_{t,i}''$ transformed target point $x_{t,i}$ w.r.t. to the predicted affine transformation and the flow field.

\textbf{Auxiliary segmentation network.} 
The majority of correspondences are localized on the foreground. 
By aggregating this information over a large set of image pairs we should be able to identify foreground objects.
Therefore, we create an auxiliary network for pixel-wise foreground classification. 
The network shares the architecture and weights of the flow network except the last layer and a sigmoid function which produces the segmentation mask. 
As our model is trained in an weakly supervised manner, we approximate the ground truth segmentation mask using neighboring pixels of the mined correspondences $M_s^{\text{corr}}$.
Similar to classification, we minimize the pixel-wise cross-entropy loss 
\begin{equation}
\begin{split}
L_{\text{mask}}= \frac{1}{h_s w_s} \sum^{h_s}_{y=1} \sum^{w_s}_{x=1}&\big( M_s^{\text{corr}}(x,y)\log (M(x,y))\\ 
&+(1-M^{\text{corr}}_{s}(x,y))\log (1-M(x,y)) \big)
\end{split}
\end{equation}
of the predicted segmentation mask $M$ and the segmentation mask extracted from our correspondences $M^{\text{corr}}_{s}$.
%
Given two images our network produces two probability maps which predict how likely each pixel belongs to the foreground. 
As postprocessing step we normalize the output of the
softmax function to the unit interval and apply the fully-connected CRF of Krähenbühl et. al \cite{FullyConnectedCRFKraehenbuehl}, which results in a binary segmentation mask.

\textbf{Final loss.}
We train the semantic flow network by taking a weighted average of the individual losses:
\begin{equation}
\begin{split}
L_{\text{total}}
=
L_{\text{aff}} 
+
\gamma L_{\text{flow}}
+
\mu L_{\text{corr}}
+
\nu L_{\text{mask}}
\end{split}
\end{equation}
where $\gamma$, $\mu$ and $\nu$ are hyperparameters.
\vspace{-0.15cm}

\section{Experiments}
%
In this section we present comparative evaluations with state-of-the-art semantic matching methods and ablation studies of our method.
\vspace{-0.15cm}
\subsection{Experimental details}
In our experiments we train on the PF-PASCAL data set \cite{Ham2016ProposalFlow}, where we neglect the subdivision into image pairs and apply the correspondence
mining algorithm on each object class separately. 
We exclude test image pairs from training.
%

\textbf{Unsupervised correspondence mining.} 
For the nearest neighbor graph we set $k=10$ and generate deep pyramid features \cite{girshick2015deformable} using conv4 features from AlexNet \cite{Krizhevsky2012ImageNetNetworks} pre-trained on ILSVRC 2012 \cite{Deng2009ImageNet:Database}.
For each image we select 80 patches and generate two matching candidates per patch.
Overall our mining procedure generates around 7k image pairs with roughly 80 correspondences per pair.

%

\textbf{Feature encoder.}
As backbone architectures we use VGG-16 \cite{vgg16} pre-trained on ILSVRC 2012 \cite{Deng2009ImageNet:Database} and truncated after the fourth layer. 
To reduce the feature dimension to 128 we append a $1\text{x}1$ convolutional layer.
Our feature encoder is trained on the mined correspondences as proxy ground truth.
For training we rescaled images to $224\text{x}224$ pixels and augment the image pairs by mirroring, randomly padding, cropping and rotating each image.
For each correspondence we extracted 60 hard negatives and
set the radius of positive correspondences to 32 pixels. 
The margin of contrastive loss is 1 and we trained 12 epochs in total.

\textbf{Semantic flow network.}
We followed the same training protocol as for the feature encoder, where we set the hyperparameters $\gamma=4$, $\mu=1$ and $\nu=1$. 
We computed the loss in forward and backward directions and determined the gradients based on their sum. The radius for generating segmentation masks from correspondences is 5 pixels. We trained the network for 40 epochs.

\def\arraystretch{1}
\setlength{\tabcolsep}{4pt}
\begin{table}[t!]
\scalebox{1.0}{
\begin{minipage}[]{0.48\linewidth}
%
%
\begin{center}
\caption{\small Influence of different dilation values on PF-PASCAL \cite{Ham2016ProposalFlow} using NN matching. 
We compare our learned mixture (MSConv Learn) with a simple averaging (MSConv Avg.) and Deformable Convolutions (DeformConv) \cite{Dai2017DeformableNetworks}.
}
\vspace{0.2cm}
\label{table:dilation-ablation}
\small
\begin{tabular}{l|ccc}
\hline
\multirow{2}{*} {Methods} & \multicolumn{3}{c}{PCK} 
\\\cline{2-4} & @$0.05$ & @$0.1$ & @$0.15$  \\
\hline
\hline
Dilation 1 & 49.0 & 68.4 & 75.6 \\
Dilation 3 & 48.4 & 73.5 & 82.7 \\
Dilation 5 & 45.9 & 73.9 & 84.8 \\
DeformConv \cite{Dai2017DeformableNetworks} & 48.7 & 65.8 & 71.8 \\
\hline
MSConv - Avg. & 48.7 & 74.5 & 83.7 \\
MSConv - Learn & \textbf{53.1} & \textbf{77.2} & \textbf{86.0} \\
\hline
\end{tabular}
\end{center}
\end{minipage}
}
\hfill
\begin{minipage}[]{0.48\linewidth}
%
%
\vspace{-0.9cm}
\begin{center}
\caption{\small Learned dilation weights 
for VGG-16 \cite{vgg16} on the PF-PASCAL benchmark \cite{Ham2016ProposalFlow}, where we averaged the multi-scale convolution weights over each layer for a compact representation.}
\vspace{0.55cm}
\label{table:dilation_weights}
\scalebox{0.95}{
\begin{tabular}{l|ccccc}
\hline
\multirow{2}{*}{Layer} & \multicolumn{5}{c}{Dilation} \\ \cline{2-6}
	& $1$ & $2$ & $3$ & $4$ & $5$ \\
\hline
\hline
Conv1 & 0.58 & 0.24 & 0.07 & 0.05 & 0.06 \\
Conv2 & 0.24 & 0.34 & 0.18 & 0.12 & 0.12 \\
Conv3 & 0.18 & 0.25 & 0.20 & 0.17 & 0.19 \\
Conv4 & 0.14 & 0.14 & 0.14 & 0.16 & 0.17 \\
\hline
\end{tabular}
}
\end{center}
\end{minipage}
\end{table}
\setlength{\tabcolsep}{1.4pt}

\subsection{Evaluation benchmarks}
%
%
\hspace{\parindent}
\textbf{PF-PASCAL benchmark \cite{Ham2016ProposalFlow}.} 
We follow the evaluation protocol of \cite{Han2017SCNet:Correspondence} and use keypoint annotations of the same test pairs, where the matching accuracy is measured using the percentage of correct keypoints (PCK) \cite{DBLP:LongZD14}. 
A keypoint is counted as transferred correctly, if its Euclidean distance to the ground truth annotation is smaller than $\alpha$, where the coordinates are normalized in the range $[0, 1]$. Please see \cite{Han2017SCNet:Correspondence} for more details about this dataset.
\textbf{PASCAL-Part benchmark \cite{Zhou2015Flowweb:Correspondences}.} 
We follow the evaluation protocol of \cite{Zhou2015Flowweb:Correspondences} and evaluate on two different tasks, which are part segment matching and keypoint matching. For part segment matching, the flow field accuracy is measured based on the transformation of part segmentation masks. 
We use the weighted intersection over union (IoU) as quantitative measure, where the weights are determined by the area of each part. 
Please see \cite{Zhou2015Flowweb:Correspondences} for more details.

\textbf{Taniai benchmark \cite{Taniai2016JointImagesb}.} 
We follow the evaluation protocol of \cite{Taniai2016JointImagesb,Rocco2017ConvolutionalMatching} and measure the flow accuracy by computing PCK densely over the whole flow field of the source object.
The misalignment threshold is set to 5 pixels after resizing images so that their larger dimension is 100 pixels. 
For the NN matching we randomly sample 1000 points on the object.
To account for different orientations in this dataset we flip the second image based on the feature loss of the flipped and unflipped version. 
Please see \cite{Taniai2016JointImagesb} for more details of this dataset.
\vspace{-0.15cm}
%
%
%
\subsection{Ablation study: Multi-scale convolution}
%
We investigate the influence of dilation levels and fusion methods.
From Tab. \ref{table:dilation-ablation} we see that by increasing the dilation stride the matching accuracy increases, although the theoretical receptive field already covers the whole image.
This is in accordance with Luo et al. \cite{Luo2017UnderstandingNetworks}, 
who found that the effective receptive field size is much smaller than the theoretical predictions. 
In Tab. \ref{table:dilation_weights} we provide the learned weights for each layer 
after training on our mined correspondences.
It shows that in the first few layers the network prefers convolutions with small dilation levels, but with higher layers this preference is reversed. 
This is reasonable, since in early layers the network learns low-level concepts 
which require a high spatial resolution while later layers tend to learn more semantic and contextual relations where a large receptive field is beneficial.
Therefore, by adjusting the dilation weight for each multi-scale convolution during the training process we obtain a clear improvement compared to a simple averaging approach.
\vspace{-0.15cm}

\bgroup
\def\arraystretch{1}
\setlength{\tabcolsep}{4pt}
\begin{table}[t!]
\begin{minipage}[t]{0.48\linewidth}
\begin{center}
\caption{\small Comparison of our correspondence mining (Ours) against ground truth data (GT) for NN matching on PF-PASCAL \cite{Ham2016ProposalFlow}. In addition we split classes and trained on one half and tested on the other half (class split).
}
\vspace{0.2cm}
\label{table:corr_mining}
\scalebox{0.95}{
\begin{tabular}{l|ccc}
\hline
\multirow{2}{*}{Methods} & \multicolumn{3}{c}{PCK} \\\cline{2-4}
& @$0.05$ & @$0.1$ & @$0.15$\\
\hline
\hline
%
%
GT & 53.1 & 78.1 & 87.4 \\
Ours & 53.1 & 77.2 & 86.0\\
\hline
GT  (class split) & 28.7 & 54.9 & 69.5 \\
Ours (class split) & 39.0 & 61.8 & 73.5 \\
%
\hline
\end{tabular}
}
\end{center}
\end{minipage}
\hfill
\begin{minipage}[t]{0.48\linewidth}
\begin{center}
\caption{\small 
Ablation study of our semantic flow network on the PF-PASCAL benchmark \cite{Ham2016ProposalFlow} using NN matching,
where we removed the affine and flow network and also disabled the feature losses $L_{\text{aff}}$ and $L_{\text{flow}}$.
}
\vspace{0.2cm}
\label{table:flow-ablation}
\scalebox{0.95}{
\begin{tabular}{l|c}
\hline
Methods & PCK@0.1  \\
\hline
\hline
Ours w/NN & 77.2  \\
\hline
Ours wo/Flow Network & 75.8  \\
Ours wo/Affine Network & 77.4  \\
Ours wo/Feature Loss & 77.0  \\
Ours & \bf{78.5} \\
\hline
\end{tabular}
}
\end{center}
\end{minipage}
\end{table}
\setlength{\tabcolsep}{1.4pt}
\egroup

\subsection{Ablation study: Correspondence mining}
%
We evaluate our correspondence mining approach by comparing NN matching accuracies 
using features trained on our mined correspondences against features trained on ground truth annotations, see Tab. \ref{table:corr_mining}.
First, we trained each method on the whole training set and evaluated on the test set using all classes, see the upper part of Tab. \ref{table:corr_mining}. It shows that by training on our mined correspondences we obtain almost identical matching accuracies without any keypoint or image pair labels. 
This is remarkable since our mining generates a lot of inconsistent image pairs with many false correspondences.
To compare the generalization capabilities of the learned features we select random splits of the dataset classes (class split) and train on one half and test on the other half. From the bottom part of Tab. \ref{table:corr_mining} we observe that our mining procedure generates feature encodings which generalize better to unseen objects.
\vspace{-0.15cm}

\subsection{Ablation study: Semantic flow network}
In Tab. \ref{table:flow-ablation} we provide an ablation study of our proposed semantic flow network. It can clearly be seen that all design choices of our basic model improved the performance. This includes the consecutive execution of an affine transformation with a flow field.
This is conceptually similar to \cite{kim2018rtns} but circumvents the expensive training of a recurrent network.
Furthermore, our experiments have shown that the feature losses $L_{\text{aff}}$ and $L_{\text{flow}}$ are beneficial since they provide an additional training signal in regions where no correspondences are available. 
\newcommand{\vheight}{0.05cm} 
\newcommand{\vwidth}{-0.01cm} 
\newcommand{\vswidth}{-0.01cm} 
%
\begin{figure*}[!t]
\centering
\begin{subfigure}[b]{.115\linewidth}
    \centering
    \includegraphics[width=.98\linewidth, height=1.35cm]{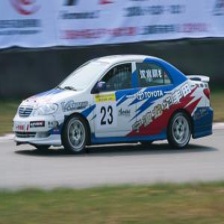} \\
	\vspace{\vheight}
	    \includegraphics[width=.98\linewidth, height=1.35cm]{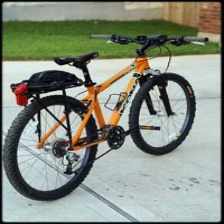}\\
    \vspace{-0.15cm}
	\caption*{\small Source} 
\end{subfigure}
\hspace{-0.15cm}
\begin{subfigure}[b]{.115\linewidth}
    \centering
    \includegraphics[width=.98\linewidth, height=1.35cm]{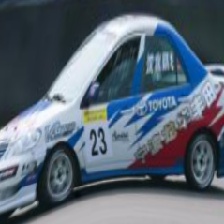} \\
	\vspace{\vheight}
	    \includegraphics[width=.98\linewidth, height=1.35cm]{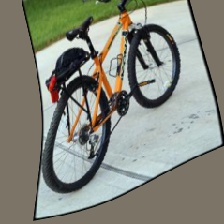}\\
    \vspace{-0.15cm}
	\caption*{\small WAlign}
\end{subfigure}
\hspace{-0.15cm}
\begin{subfigure}[b]{.115\linewidth}
    \centering
    \includegraphics[width=.98\linewidth, height=1.35cm]{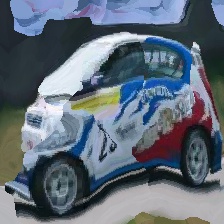} \\
	\vspace{\vheight}
	    \includegraphics[width=.98\linewidth, height=1.35cm]{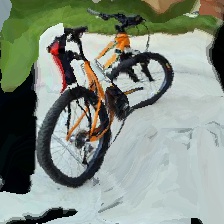}\\
    \vspace{-0.15cm}
	\caption*{\small Ours NN}
\end{subfigure}
\hspace{-0.15cm}
\begin{subfigure}[b]{.115\linewidth}
    \centering
    \includegraphics[width=.98\linewidth, height=1.35cm]{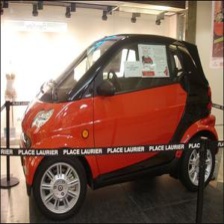} \\
	\vspace{\vheight}
	    \includegraphics[width=.98\linewidth, height=1.35cm]{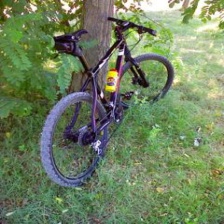}\\
    \vspace{-0.15cm}
    \caption*{\small Target}
\end{subfigure}
\hspace{\vwidth}
\begin{subfigure}[b]{.115\linewidth}
    \centering
    \includegraphics[width=.98\linewidth, height=1.35cm]{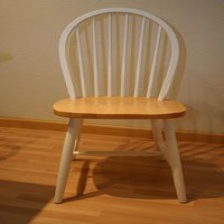}\\
	\vspace{\vheight}
    \includegraphics[width=.98\linewidth, height=1.35cm]{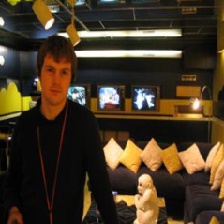}\\
    \vspace{-0.15cm}
	\caption*{\small Source} 
\end{subfigure}
\hspace{-0.15cm}
\begin{subfigure}[b]{.115\linewidth}
    \centering
    \includegraphics[width=.98\linewidth, height=1.35cm]{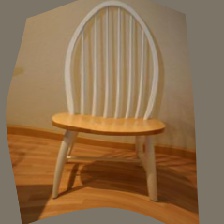}\\
	\vspace{\vheight}
    \includegraphics[width=.98\linewidth, height=1.35cm]{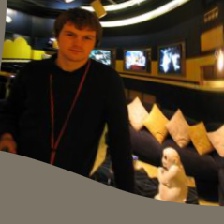}\\
    \vspace{-0.15cm}
	\caption*{\small WAlign}
\end{subfigure}
\hspace{-0.15cm}
\begin{subfigure}[b]{.115\linewidth}
    \centering
    \includegraphics[width=.98\linewidth, height=1.35cm]{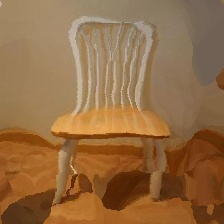}\\
	\vspace{\vheight}
    \includegraphics[width=.98\linewidth, height=1.35cm]{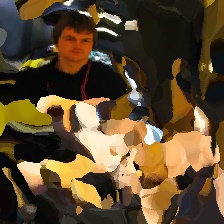}\\
    \vspace{-0.15cm}
	\caption*{\small Ours NN}
\end{subfigure}
\hspace{-0.15cm}
\begin{subfigure}[b]{.115\linewidth}
    \centering
    \includegraphics[width=.98\linewidth, height=1.35cm]{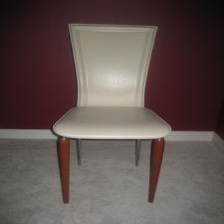}\\
	\vspace{\vheight}
    \includegraphics[width=.98\linewidth, height=1.35cm]{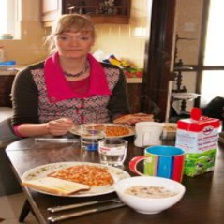}\\
    \vspace{-0.15cm}
    \caption*{\small Target}
\end{subfigure}
\vspace{-0.14cm}
\caption[]{ \small 
Qualitative examples on the PF-PASCAL dataset \cite{Han2017SCNet:Correspondence},
which shows a source, target and the warped images using NN matching with our features (Ours NN) and the weak alignment method (WAlign) \cite{Rocco2017ConvolutionalMatching}.} 
\label{fig:resultsPfPascal1}
\end{figure*}

%
\begin{figure*}[!t]
\centering
\begin{subfigure}[b]{.137\linewidth}
    \centering
    \includegraphics[width=.98\linewidth, height=1.35cm]{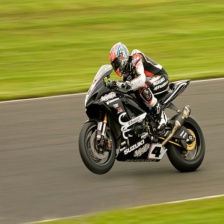} \\
	\vspace{0.05cm}
    \includegraphics[width=.98\linewidth, height=1.35cm]{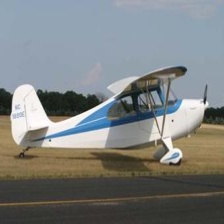} \\
    \vspace{-0.15cm}
	\caption*{\small Source} 
\end{subfigure}
\hspace{0.05cm}
\begin{subfigure}[b]{.137\linewidth}
    \centering
    \includegraphics[width=.98\linewidth, height=1.35cm]{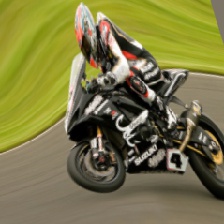} \\
	\vspace{0.05cm}
    \includegraphics[width=.98\linewidth, height=1.35cm]{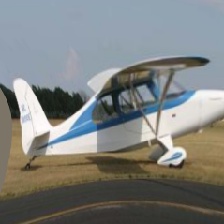} \\
    \vspace{-0.15cm}
	\caption*{\small WAlign}
\end{subfigure}
\hspace{0.05cm}
\begin{subfigure}[b]{.137\linewidth}
    \centering
    \includegraphics[width=.98\linewidth, height=1.35cm]{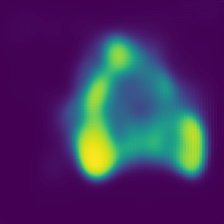} \\
	\vspace{0.05cm}
    \includegraphics[width=.98\linewidth, height=1.35cm]{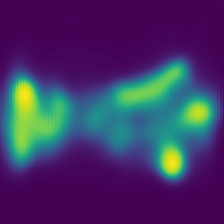} \\
    \vspace{-0.15cm}
	\caption*{\small Ours Prob.}
\end{subfigure}
\hspace{-0.15cm}
\begin{subfigure}[b]{.137\linewidth}
    \centering
    \includegraphics[width=.98\linewidth, height=1.35cm]{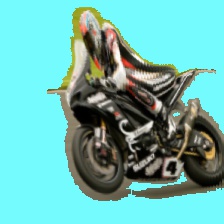} \\
	\vspace{0.05cm}
    \includegraphics[width=.98\linewidth, height=1.35cm]{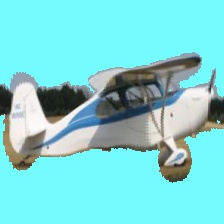} \\
    \vspace{-0.15cm}
    \caption*{\small Ours}
\end{subfigure}
\hspace{0.05cm}
\begin{subfigure}[b]{.137\linewidth}
    \centering
    \includegraphics[width=.98\linewidth, height=1.35cm]{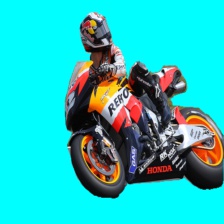} \\
	\vspace{0.05cm}
    \includegraphics[width=.98\linewidth, height=1.35cm]{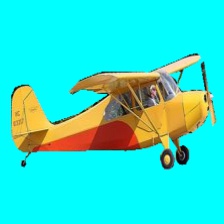} \\
    \vspace{-0.15cm}
	\caption*{\small Target Seg.} 
\end{subfigure}
\hspace{-0.15cm}
\begin{subfigure}[b]{.137\linewidth}
    \centering
    \includegraphics[width=.98\linewidth, height=1.35cm]{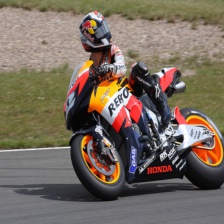} \\
	\vspace{0.05cm}
    \includegraphics[width=.98\linewidth, height=1.35cm]{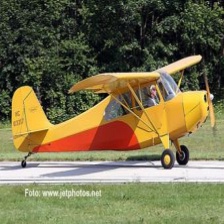} \\
    \vspace{-0.15cm}
	\caption*{\small Target}
\end{subfigure}

\vspace{-0.15cm}
\caption[]{ \small 
Qualitative examples on the Taniai dataset \cite{Taniai2016JointImagesb},
which shows a source image, the warped images with the weak alignment method (WAlign) \cite{Rocco2017ConvolutionalMatching}, the probability map of our segmentation network (Ours Prob.), the transformation of our semantic flow network with segmentation (Ours), the ground truth segmentation of the target (Target Seg.) and the target image.
}
\label{fig:resultsTaniai}
\end{figure*}

\begin{figure}[!t]
\centering
\begin{minipage}{.15\linewidth}
    \centering
	\includegraphics[trim=10 23 15 40, clip, width=.97\linewidth, height=1.4cm]{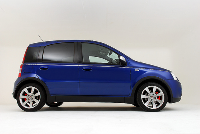}
    {\small Source}
\end{minipage}%
\begin{minipage}{.15\linewidth}
    \centering
	\includegraphics[width=.97\linewidth, height=1.4cm]{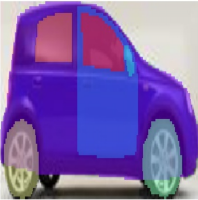}
    {\small Source Parts}
\end{minipage}%
\begin{minipage}{.15\linewidth}
    \centering
	\includegraphics[width=.97\linewidth, height=1.4cm]{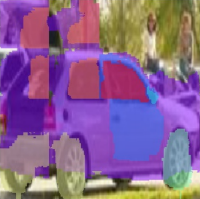}
    {\small DSP}
\end{minipage}%
\begin{minipage}{.15\linewidth}
    \centering
	\includegraphics[width=.97\linewidth, height=1.4cm]{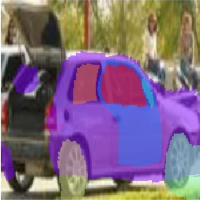}
    {\small Ours}
\end{minipage}%
\begin{minipage}{.15\linewidth}
    \centering
	\includegraphics[width=.97\linewidth, height=1.4cm]{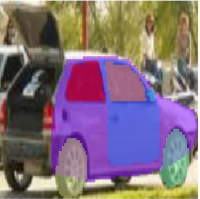}
    {\small Target Parts}
\end{minipage}%
\begin{minipage}{.15\linewidth}
    \centering
	\includegraphics[trim=20 20 47 36, clip, width=.97\linewidth, height=1.4cm]{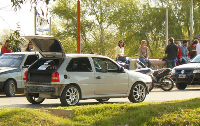}
    {\small Target}
\end{minipage}%
\vspace{-0.15cm}
\caption{\small Visualization of part segment transformations on the PASCAL Parts dataset \cite{Chen2014DetectParts}, which shows a source image,  the color-coded part masks, part segment transformations with DSP \cite{Kim2013DeformableCorrespondences} and with our semantic flow network (Ours), target masks and the target.}
\label{fig:pascalpart}
\end{figure}

\begin{table}
\begin{center}
\caption{\small 
Evaluation results on the PF-PASCAL \cite{Ham2016ProposalFlow}, Taniai \cite{Taniai2016JointImagesb} and PASCAL Parts \cite{Ham2016ProposalFlow} dataset.
The amount of supervision ranges from none (-), image class labels (classL), image pairs (imgPairs), bounding boxes (bbx) up to full keypoints annotations (keyPts). 
}
\vspace{0.2cm}
\label{table:all-results}
\scalebox{0.95}{
\begin{tabular}{l|c|c|c|c|c:c:c}
\hline
\multirow{2}{*} {Methods} & 
\multirow{2}{*} {Feat. Extr.} & 
\multirow{2}{*} {Superv.} & 
\multicolumn{1}{c|}{PF-PASCAL}  
& \multicolumn{1}{c|}{\ Taniai \quad}
& \multicolumn{3}{c}{PASCAL Parts}
\\\cline{4-8}  &  & & PCK@$0.1$ & Avg. & \ IoU \quad &  \quad @$0.05$ \quad & \quad @$0.1$ \quad   \\

\hline
\hline
DSP \cite{Kim2013DeformableCorrespondences} & SIFT & -- & 37 & 44.5  & 0.39 & 17 & --  \\ 
\hdashline
PF \cite{Ham2016ProposalFlow} & HOG & -- & 62.5 & 65.7 & 0.41 & 17 & 36  \\ 
\hdashline
DCTM \cite{Kim2017DCTM:Flow} & VGG-16 & -- & 69.6 & 74.0 & 0.48 & 32 & 50  \\ 
\hdashline
UCN-ST \cite{Choy2016UniversalNetwork} & Inception-v1 & keyPts & 55.6 & 67.9 & -- & 26 & 44 \\
\hdashline
FCSS-LOM \cite{Kim2017FCSS:Correspondence}& VGG-16 & bbox & 68.9 & 66.8 & 0.44 & 28 & 47  \\ \hdashline
SCNet-AG+ \cite{Han2017SCNet:Correspondence}& VGG-16 & keyPts & 72.2 &  61.9 & 0.48 & 18 & --   \\ \hdashline
CNNGeo \cite{Rocco2017End-to-endAlignment} & VGG-16 & synth & 62.6 & 67.3 & 0.51 & 22.1 & 46.6 \\ \hdashline
CNNGeo \cite{Rocco2017End-to-endAlignment} & ResNet-101 & synth & 69.5 & 73.5 & 0.54 & 25.8 & 49.5 \\ \hdashline
WAlign \cite{Rocco2017ConvolutionalMatching} & ResNet-101 & \makecell{synth \& \\ imgPairs} & 74.8 & 73.7 & 0.55  & 28.4 & 53.2 \\ \hdashline
RTNs \cite{kim2018rtns} & VGG-16 & \ imgPairs \ & 74.3 & 73.2 & --  & -- & --\\ \hdashline

RTNs \cite{kim2018rtns} & ResNet-101 & imgPairs & 75.9 & 77.2 & --  & -- & --\\ \hline
Ours w/NN & VGG-16 & classL &  77.2 &  75.7 & -- & \bf{37.2} & 54.3 \\
\hdashline
Ours & VGG-16 & classL &  \bf{78.5} & \bf{79.7} & \bf{0.57} & 34.7 & \bf{54.5} \\
\hline
\end{tabular}
}
\end{center}
\end{table}

 
%
%
%
%
%
\subsection{Results on benchmarks}
\vspace{-0.05cm}
%
\hspace{\parindent}
\textbf{Qualitative results.}
Fig. \ref{fig:resultsPfPascal1} shows qualitative results on the PF-PASCAL dataset \cite{Ham2016ProposalFlow} with our features using nearest neighbor matching.
It can clearly be seen that our matching provides more natural warps and is able to capture difficult non-linear deformations.
Please notice the difficulty of this task 
as demonstrated in Fig. \ref{fig:first_teaser}. 
Examples on the Taniai dataset \cite{Taniai2016JointImagesb} can be seen in Fig. \ref{fig:resultsTaniai}.
In contrast to previous deep learning based approaches our network is also capable to predict foreground segmentation masks.
In Fig. \ref{fig:pascalpart} we provide an example of part matchings on the PASCAL Parts dataset \cite{Chen2014DetectParts}.

\textbf{Quantitative results.}
In Tab. \ref{table:all-results} we present quantitative comparisons. 
We observe that nearest neighbor matching with our features outperform all existing semantic matching methods on the PF-PASCAL dataset \cite{Ham2016ProposalFlow}, including supervised methods [12]. 
Furthermore our semantic flow network outperforms current benchmarks on PF-PASCAL \cite{Ham2016ProposalFlow}, Taniai \cite{Taniai2016JointImagesb} and PASCAL Parts \cite{Chen2014DetectParts}. 
We evaluate our auxiliary segmentation network also for the task of co-segmentation and obtain superior results compared to state-of-the-art on the Taniai dataset \cite{Taniai2016JointImagesb}, see Tab. \ref{table:co-segmentation}.
\begin{table}[t]
\begin{center}
\small
\caption{\small Evaluation of our segmentation network for co-segmentation on the Tania benchmark \cite{Taniai2016JointImagesb}.}
\vspace{0.2cm}
\scalebox{0.95}{
\begin{tabular}{c||c|c|c|c}
\hline
\phantom{tt} Methods \phantom{tt} & Joulin et al. \cite{CoSegmentationJoulin} & Faktor et al. \cite{CoSegmentationFaktor} &  Tania et al. \cite{Taniai2016JointImagesb} & \phantom{test} Ours  \phantom{test} \\
\hline
IoU & 0.39 & 0.58 & 0.64 & \bf{0.72} \\
\hline
\end{tabular}
}
\label{table:co-segmentation}
\end{center}
\end{table} 
%
%
%
%
%

\section{Conclusion}
\vspace{-0.15cm}
We have presented a new convolutional layer for semantic matching which learns highly context aware feature representations with a high localization accuracy.
It clearly outperforms other methods like dilated and deformable convolutions.
To alleviate tedious manual labeling, we have introduced a correspondence mining procedure which requires only class labels and is more effective than training on ground truth keypoints. 
Therefore, we require less supervision than the current best methods.
Our semantic flow network produces more natural warps compared to parametric transformation based models and is also capable to provide reliable foreground segmentation masks at the same time.
The presented approach demonstrated significant improvements on several challenging benchmark datasets for semantic matching and joint co-segmentation. 
In future work we want to learn more powerful features by applying our approach on a large-scale dataset such as ImageNet.

\bibliographystyle{splncs04}
\bibliography{arxive_2019}

\newpage

\section*{A. Supplementary Material}

\section*{A.1 Multi-scale convolution}
%
As described in Sect. 3.2 of the main paper, we replace all convolutional layers of the feature extractor with the proposed multi-scale convolutions.
In the following we provide an additional ablation study and visualization the effect on the receptive field.

\subsection*{A.1.1 Evaluation using AlexNet \cite{Krizhevsky2012ImageNetNetworks}}
To demonstrate that the proposed multi-scale convolution generalizes to other network architectures, we provide an ablation study analogously to Sect. 4.3 of the main paper for AlexNet \cite{Krizhevsky2012ImageNetNetworks}, see Tab. \ref{table:dilation-ablation},
and provide the learned dilations weights in Tab. \ref{table:dilation_weights}.
Overall, we observe a similar behaviour as we already described in the main paper. The multi-scale convolutions clearly improves the matching accuracies compared to a simple averaging approach.
\vspace{-0.05cm}

\def\arraystretch{1}
\setlength{\tabcolsep}{4pt}
\begin{table}[h!]
\scalebox{1.0}{
\begin{minipage}[]{0.48\linewidth}
%
%
\begin{center}
\caption{\small Influence of different dilation values on PF-PASCAL \cite{Ham2016ProposalFlow} using NN matching. 
We compare our learned mixture (MSConv Learn) with a simple averaging (MSConv Avg.).
}
\vspace{0.2cm}
\label{table:dilation-ablation}
\small
\begin{tabular}{l|ccc}
\hline
\multirow{2}{*} {Methods} & \multicolumn{3}{c}{PCK, AlexNet \cite{Krizhevsky2012ImageNetNetworks}  } 
\\\cline{2-4} & @$0.05$ & @$0.1$ & @$0.15$  \\
\hline
\hline
%
Dilation 1 & 43.2 & 64.1 & 72.3 \\
Dilation 3 & 42.9 & 70.1 & 81.7 \\
Dilation 5 & 34.8 & 66.9 & 81.1 \\
\hline
MSConv - Avg. & 44.4 & 68.1 & 79.0 \\
MSConv - Learn & \textbf{49.1} & \textbf{74.2} & \textbf{84.0} \\
\hline
\end{tabular}
\end{center}
\end{minipage}
}
\hfill
\begin{minipage}[]{0.48\linewidth}
%
%
\vspace{-0.5cm}
\begin{center}
\caption{\small Learned dilation weights 
for AlexNet \cite{Krizhevsky2012ImageNetNetworks} on the PF-PASCAL dataset \cite{Ham2016ProposalFlow}.}
\vspace{0.9cm}
\label{table:dilation_weights}
\scalebox{0.95}{
\begin{tabular}{l|ccccc}
\hline
\multirow{2}{*}{Layer} & \multicolumn{5}{c}{Dilation, AlexNet \cite{Krizhevsky2012ImageNetNetworks} } \\ \cline{2-6}
	& $1$ & $2$ & $3$ & $4$ & $5$ \\
\hline
\hline
%
%

Conv1 & 0.69 & 0.20 & 0.05 & 0.03 & 0.03 \\
Conv2 & 0.38 & 0.17 & 0.14 & 0.14 & 0.17 \\
Conv3 & 0.12 & 0.17 & 0.24 & 0.27 & 0.19 \\
Conv4 & 0.16 & 0.19 & 0.21 & 0.22 & 0.22 \\
\hline
\end{tabular}
}
\end{center}
\end{minipage}
\end{table}
\setlength{\tabcolsep}{1.4pt}
\vspace{-0.05cm}
%

%
%
%
%
%
%
%
%
\subsection*{A.1.2 Visualization of the effective receptive field}
\begin{figure*}[h!]
\begin{center}
\begin{subfigure}[b]{.18\linewidth}
    \centering
	\includegraphics[width=.95\linewidth, height=1.8cm]{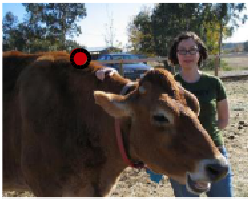}\\
	\includegraphics[width=.95\linewidth, height=1.8cm]{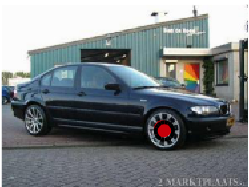} \\
    \vspace{-0.1cm}
	\captionsetup{justification=centering}
	\subcaption*{Source \\ \phantom{Dilation 1}} 
\end{subfigure}
\hspace{-0.125cm}
\begin{subfigure}[b]{.18\linewidth}
    \centering
	\includegraphics[width=.95\linewidth, height=1.8cm]{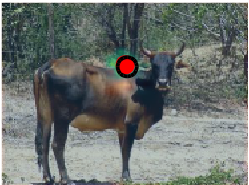} \\
    \includegraphics[width=.95\linewidth, height=1.8cm]{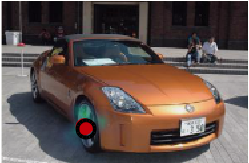}\\
    \vspace{-0.1cm}
	\captionsetup{justification=centering}
	\caption*{Target \\ Dilation 1} 
\end{subfigure}
\hspace{-0.15cm}
\begin{subfigure}[b]{.18\linewidth}
    \centering
	\includegraphics[width=.95\linewidth, height=1.8cm]{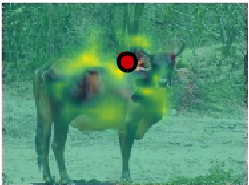} \\
    \includegraphics[width=.95\linewidth, height=1.8cm]{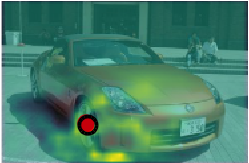}  \\
    \vspace{-0.1cm}
	\captionsetup{justification=centering}
	\caption*{Target \\ MSConv}
\end{subfigure}
\vspace{-0.125cm}
\caption[]{\small Coverage experiment. 
For both examples, the first image shows a query (Source), followed by the target (Target) with the effective receptive field using dilation 1 (Dilation 1) and the multi-scale convolutions (MSConv), respectively. 
The corresponding keypoints used for measuring the similarity are marked with red dots in each image.
}
\label{fig:coverage_experiment}
\end{center}
\end{figure*}

In the following we visualize the increased effective receptive field of our multi-scale convolution based feature extractor as we argued in the main paper.
Inspired by Zeiler et al. \cite{Zeiler2014Visualizing2013}, we make the following coverage visualization.
Given a target and source image we fix two points and measure their feature similarity depending on the position of a grey square, which slides over the target image and visualize the change of the similarity score as heatmap in the target image. see Fig. \ref{fig:coverage_experiment}.
This allows us to visualize the regions in the image which are important for the feature similarity. 
For the standard network configuration with dilation 1, we observe that the region which influences the similarity occupies only a small fraction of the theoretical receptive field (second column) and significantly increases by our multi-scale convolutions (third column). 
%
%

\section*{A.2 Correspondence mining}

\subsection*{A.2.1 Examples of mined correspondences}
We visualize results of the corresponding mining approach, see Fig. \ref{fig:sparseMatches}.
In most cases it delivers consistent image pairs with accurate correspondences.
In situations where image pairs are inconsistent or the discovered main geometry is not in agreement with the object depicted in both images the introduced outlier class supresses a lot of wrong correspondences.
The resulting proxy segmentation masks are very sparse and inaccurate. 
However, over a large set of examples the segmentation network learns to generate accurate foreground predictions. 

\subsection*{A.2.2 Generalization to other datasets}
Despite the trend of state-of-the-art methods \cite{Rocco2017End-to-endAlignment,kim2018rtns} to train on PF-PASCAL \cite{Ham2016ProposalFlow}
some previous methods 
\cite{Kim2017FCSS:Correspondence,kim2018fcss}
utilized the Caltech-101 dataset \cite{Fei-Fei2006:One-ShotLearning}.
%
%
To provide a fair comparison and to show that our feature learning approach generalizes to other datasets, we also train our method on the
Caltech-101 dataset \cite{Fei-Fei2006:One-ShotLearning} and report matching accuracies on the PF-PASCAL benchmark \cite{Ham2016ProposalFlow}, see Tab. \ref{table:generalization}.
%
%
\begin{table}[t!]
\begin{center}
\caption{\small Matching accuracy on the PF-PASCAL benchmark \cite{Ham2016ProposalFlow}, where we compare our performance for training on different datasets.   
}
\vspace{0.2cm}
\label{table:generalization}
\scalebox{0.95}{
\begin{tabular}{l|c|ccc}
\hline
\multirow{2}{*}{Method} & \multirow{2}{*}{Training Dataset} & \multicolumn{3}{c}{PCK} \\\cline{3-5}
& & @$0.05$ & @$0.1$ & @$0.15$\\
\hline
\hline
FCSS-CAT \cite{kim2018fcss} & Caltech-101 \cite{Fei-Fei2006:One-ShotLearning} & 33.6 &  68.9 & 79.2 \\
Ours & Caltech-101 \cite{Fei-Fei2006:One-ShotLearning} & 40.0 & 71.1 & 84.6 \\
\hline
Ours & PF-PASCAL \cite{Ham2016ProposalFlow} & 52.7 & 78.5 & 88.1 \\
\hline
\end{tabular}
}
\end{center}
\end{table}

\noindent
Since we train on another dataset than PF-PASCAL \cite{Ham2016ProposalFlow} the performance drops significantly, nevertheless, we outperform the current best method, which is trained on Caltech-101 \cite{Fei-Fei2006:One-ShotLearning}.

\section*{A.3 Evaluation on benchmarks}

\subsection*{A.3.1 Comparison with PARN \cite{jeon2018parn}}
\label{sec:pf-pascal-benchmark}
The recent publication of Jeaon et al. \cite{jeon2018parn}
follows another evaluation protocol for the PF-PASCAL benchmark \cite{Han2017SCNet:Correspondence} than we use in the main paper. 
In order to compare with their method we provide our matching accuracies with this modified definition of percentage of correct keypoints (PCK).
They define a keypoint to be correctly transformed if the distance between warped and ground-truth keypoint is smaller than $\alpha \cdot \max(h,w)$ where $\alpha \in [0,1]$ and $h$, $w$ are the height and width of the bounding box in the target image.
This leads to a smaller reference length and overall smaller PCK values.
From Tab. \ref{table:pfpascalresults} we can see that we clearly outperform Jeon et al. \cite{jeon2018parn}, although they use more supervision in terms of segmentation masks.
%
%

\bgroup
\begin{table}[t!]
\begin{center}
\caption{\small Matching accuracy on the PF-PASCAL dataset \cite{Han2017SCNet:Correspondence}, where we use the definition of PCK from Joen et al. \cite{jeon2018parn}.
The amount of required supervision (Superv.) ranges from no supervision (-), image class labels (classL), 
segmentation masks (segm) up to full keypoint annotations (keyPts). 
}
\vspace{0.2cm}
\label{table:pfpascalresults}
\scalebox{0.95}{
\begin{tabular}{l|c|ccc}
\hline
\multirow{2}{*}{Methods} &  \multirow{2}{*}{Superv.} & \multicolumn{3}{c}{PCK} \\\cline{3-5}
& & @$0.05$ & @$0.1$ & @$0.15$ \\
\hline
\hline
PF \cite{Ham2016ProposalFlow} & -- & 19.2 & 33.4 & 49.2 \\
SCNet \cite{Han2017SCNet:Correspondence} & keyPts & 26.0 & 48.2 & 65.8 \\
PARN \cite{jeon2018parn} & segm & 26.8 & 49.1 & 66.2 \\
\hline
Ours w/NN  & classL & \textbf{42.9} & 66.5 & 78.8  \\
Ours & classL & 42.8 & \textbf{67.8} & \textbf{80.2} \\ 
\hline
\end{tabular}
}
\end{center}
\end{table}

\noindent

\subsection*{A.3.2 Further qualitative results}
In this section, we provide additional qualitative results of nearest neighbor matching with our features and joint registration and segmentation using our semantic flow network on the PF-PASCAL \cite{Ham2016ProposalFlow} and the Taniai \cite{Taniai2016JointImagesb} dataset, see Figures \ref{fig:resultsPfPascal2} and \ref{fig:resultsTaniai1},
respectively. 
The results support the statement that our algorithm is capable to warp images pixel-wise
accurately across objects despite heavy intra-class variation, view-point changes and background
clutter.
In addition the segmentation network is capable to predict accurately the foreground in both images.

%
\begin{figure*}[!h]
\centering
\begin{subfigure}[b]{.44\linewidth}
    \centering
    \includegraphics[width=.97\linewidth, height=2.2cm]{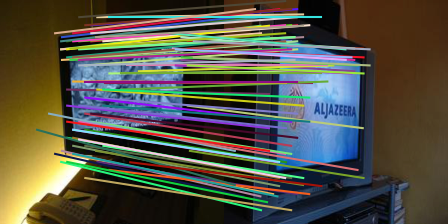} \\
    \includegraphics[width=.97\linewidth, height=2.2cm]{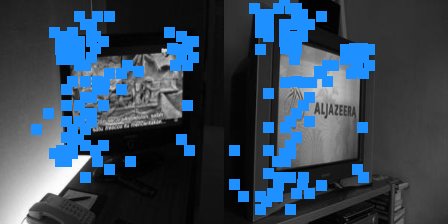} \\
	\vspace{0.25cm}
    \includegraphics[width=.97\linewidth, height=2.2cm]{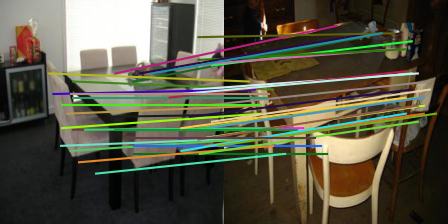} \\
    \includegraphics[width=.97\linewidth, height=2.2cm]{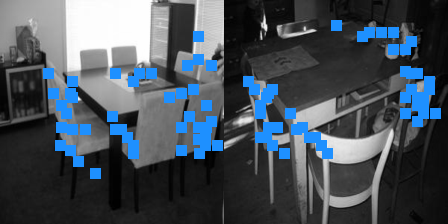} \\
	\vspace{0.25cm}
    \includegraphics[width=.97\linewidth, height=2.2cm]{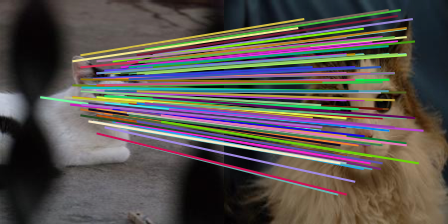} \\
    \includegraphics[width=.97\linewidth, height=2.2cm]{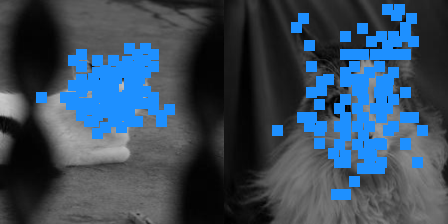} \\
	\vspace{0.25cm}
    \includegraphics[width=.97\linewidth, height=2.2cm]{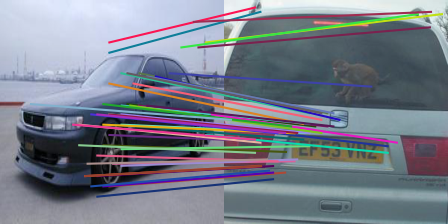} \\
    \includegraphics[width=.97\linewidth, height=2.2cm]{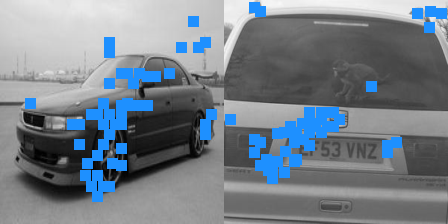} 
\end{subfigure}
\hspace{-0.15cm}
\begin{subfigure}[b]{.44\linewidth}
    \centering
    \includegraphics[width=.97\linewidth, height=2.2cm]{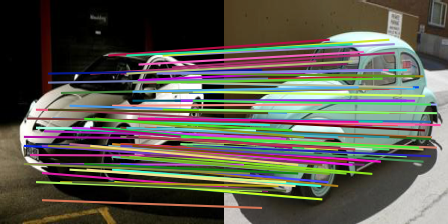} \\
    \includegraphics[width=.97\linewidth, height=2.2cm]{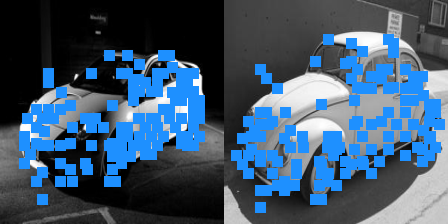} \\
	\vspace{0.25cm}
	\includegraphics[width=.97\linewidth, height=2.2cm]{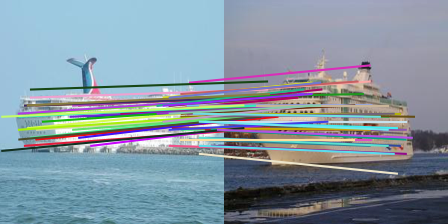} \\
    \includegraphics[width=.97\linewidth, height=2.2cm]{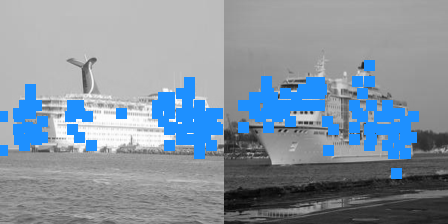} \\
	\vspace{0.25cm}
    \includegraphics[width=.97\linewidth, height=2.2cm]{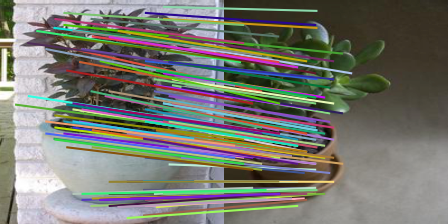} \\
    \includegraphics[width=.97\linewidth, height=2.2cm]{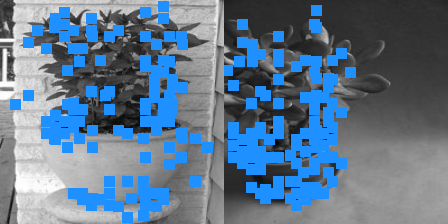} \\
	\vspace{0.25cm}
    \includegraphics[width=.97\linewidth, height=2.2cm]{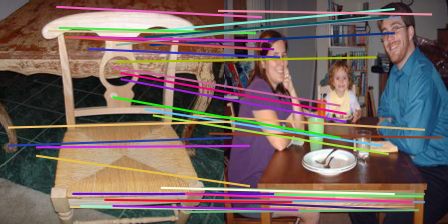} \\
    \includegraphics[width=.97\linewidth, height=2.2cm]{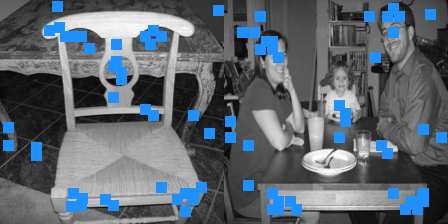} \\
\end{subfigure}
\caption[]{ \small 
Visualization of mined correspondences.
We show the sparse correspondences and the resulting proxy segmentation masks (blue squares on top of the grey scaled image) for all examples.
In the last row we show failure cases, 
where the image pairs are inconsistent, i.e. showing different objects or the same object from different viewpoints. 
}
\label{fig:sparseMatches}
\end{figure*}

%
\begin{figure*}[!t]
\centering
\begin{subfigure}[b]{.21\linewidth}
    \centering
    \includegraphics[width=.98\linewidth, height=2.2cm]{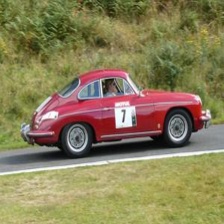} \\
	\vspace{0.05cm}
	\includegraphics[width=.98\linewidth,height=2.2cm]{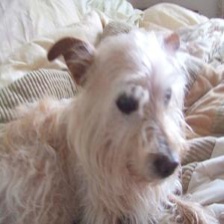}\\
	\vspace{0.05cm}
    \includegraphics[width=.98\linewidth, height=2.2cm]{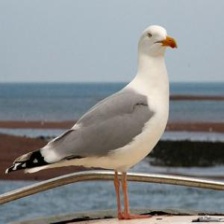}\\
	\vspace{0.05cm}
	 \includegraphics[width=.98\linewidth, height=2.2cm]{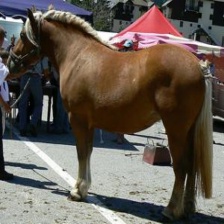}\\
	\vspace{0.05cm}
	\includegraphics[width=.98\linewidth, height=2.2cm]{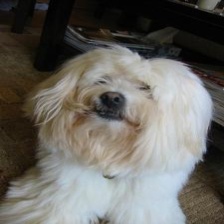}\\ \vspace{0.05cm}
	    \includegraphics[width=.98\linewidth, height=2.2cm]{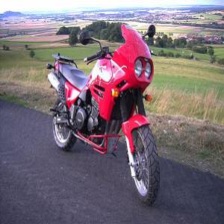}\\
	\vspace{0.05cm}
	    \includegraphics[width=.98\linewidth, height=2.2cm]{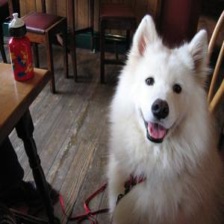}\\
	\vspace{0.05cm}
    \includegraphics[width=.98\linewidth, height=2.2cm]{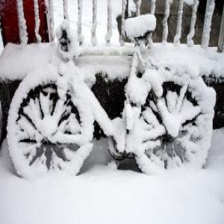}\\
    \vspace{-0.15cm}
	\caption*{\small Source} 
\end{subfigure}
\hspace{-0.15cm}
\begin{subfigure}[b]{.21\linewidth}
    \centering
    \includegraphics[width=.98\linewidth, height=2.2cm]{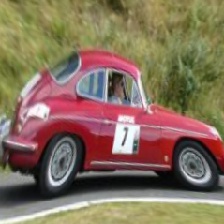} \\
	\vspace{0.05cm}
	\includegraphics[width=.98\linewidth,height=2.2cm]{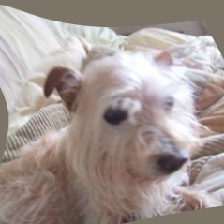}\\
	\vspace{0.05cm}
    \includegraphics[width=.98\linewidth, height=2.2cm]{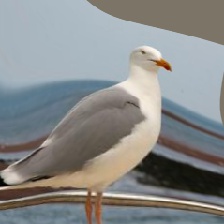}\\
	\vspace{0.05cm}
	\includegraphics[width=.98\linewidth, height=2.2cm]{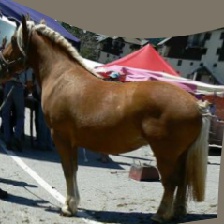}\\
	\vspace{0.05cm}
	\includegraphics[width=.98\linewidth, height=2.2cm]{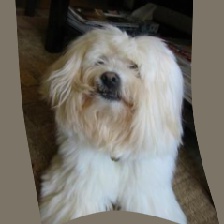}\\
	\vspace{0.05cm}
	\includegraphics[width=.98\linewidth, height=2.2cm]{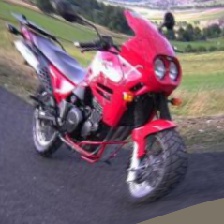}\\
	\vspace{0.05cm}
	\includegraphics[width=.98\linewidth, height=2.2cm]{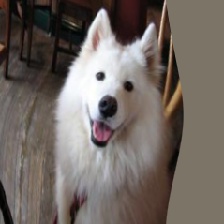}\\
	\vspace{0.05cm}
    \includegraphics[width=.98\linewidth, height=2.2cm]{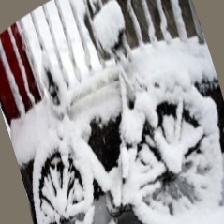}\\
    \vspace{-0.15cm}
	\caption*{\small WAlign}
\end{subfigure}
\hspace{-0.15cm}
\begin{subfigure}[b]{.21\linewidth}
    \centering
    \includegraphics[width=.98\linewidth, height=2.2cm]{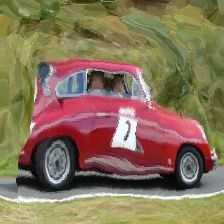} \\
	\vspace{0.05cm}
	\includegraphics[width=.98\linewidth,height=2.2cm]{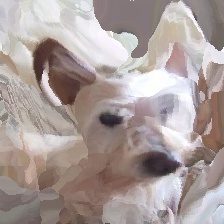}\\
	\vspace{0.05cm}
    \includegraphics[width=.98\linewidth,height=2.2cm]{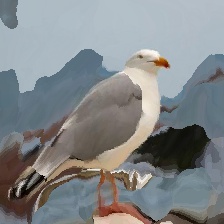}\\
	\vspace{0.05cm}
    \includegraphics[width=.98\linewidth,height=2.2cm]{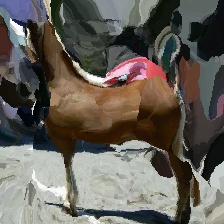}\\
	\vspace{0.05cm}
    \includegraphics[width=.98\linewidth,height=2.2cm]{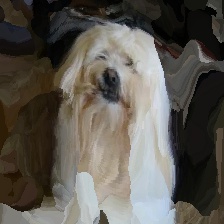}\\
	\vspace{0.05cm}
    \includegraphics[width=.98\linewidth,height=2.2cm]{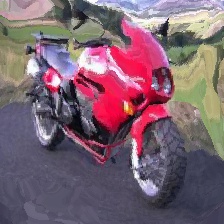}\\
	\vspace{0.05cm}
    \includegraphics[width=.98\linewidth,height=2.2cm]{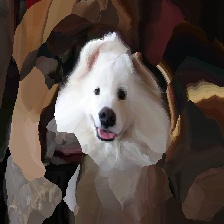}\\
	\vspace{0.05cm}
    \includegraphics[width=.98\linewidth,height=2.2cm]{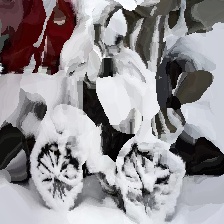}\\
    \vspace{-0.15cm}
	\caption*{\small Ours NN}
\end{subfigure}
\hspace{-0.15cm}
\begin{subfigure}[b]{.21\linewidth}
    \centering
    \includegraphics[width=.98\linewidth, height=2.2cm]{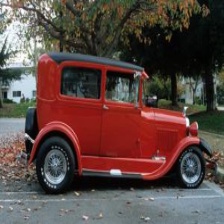} \\
	\vspace{0.05cm}
	\includegraphics[width=.98\linewidth, height=2.2cm]{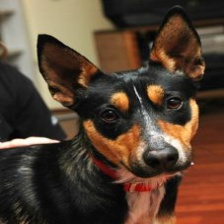}\\
	\vspace{0.05cm}
    \includegraphics[width=.98\linewidth, height=2.2cm]{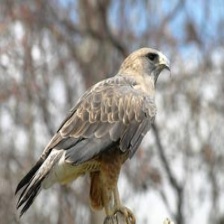}\\
	\vspace{0.05cm}
    \includegraphics[width=.98\linewidth, height=2.2cm]{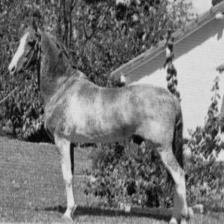}\\
	\vspace{0.05cm}
    \includegraphics[width=.98\linewidth, height=2.2cm]{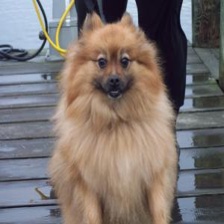}\\
	\vspace{0.05cm}
    \includegraphics[width=.98\linewidth, height=2.2cm]{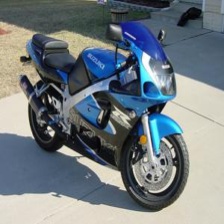}\\
	\vspace{0.05cm}
    \includegraphics[width=.98\linewidth, height=2.2cm]{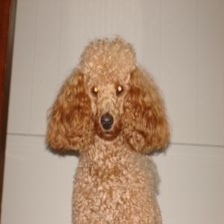}\\
	\vspace{0.05cm}
    \includegraphics[width=.98\linewidth, height=2.2cm]{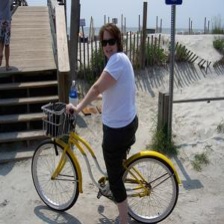}\\
    \vspace{-0.15cm}
    \caption*{\small Target}
\end{subfigure}
\vspace{-0.14cm}
\caption[]{ \small 
Qualitative examples on the PF-PASCAL dataset \cite{Han2017SCNet:Correspondence},
which shows a source, target and the warped images using NN matching with our features (Ours NN) and the weak alignment method  of Rocco et al. (WAlign) \cite{Rocco2017ConvolutionalMatching}.} 
\label{fig:resultsPfPascal2}
\end{figure*}

%
\begin{figure*}[!t]
\centering
\begin{subfigure}[b]{.18\linewidth}
    \centering
    \includegraphics[width=.98\linewidth, height=2.0cm]{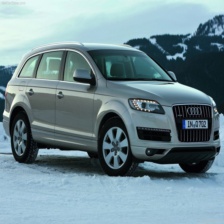} \\
    \includegraphics[width=.98\linewidth, height=2.0cm]{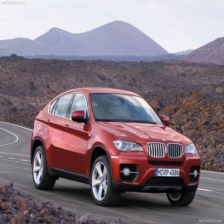} \\
	\vspace{0.15cm}
    \includegraphics[width=.98\linewidth, height=2.0cm]{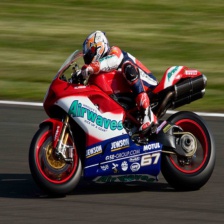} \\
    \includegraphics[width=.98\linewidth, height=2.0cm]{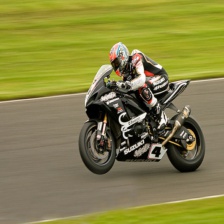} \\
	\vspace{0.15cm}
    \includegraphics[width=.98\linewidth, height=2.0cm]{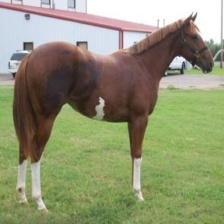} \\
    \includegraphics[width=.98\linewidth, height=2.0cm]{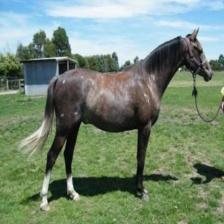} \\
	\vspace{0.15cm}
	\includegraphics[width=.98\linewidth, height=2.0cm]{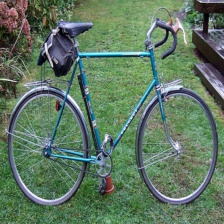} \\
    \includegraphics[width=.98\linewidth, height=2.0cm]{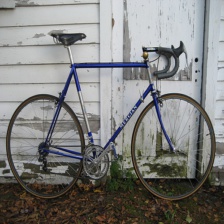} \\
    \vspace{-0.15cm}
	\caption*{\small Source} 
\end{subfigure}
\hspace{-0.15cm}
\begin{subfigure}[b]{.18\linewidth}
    \centering
    \includegraphics[width=.98\linewidth, height=2.0cm]{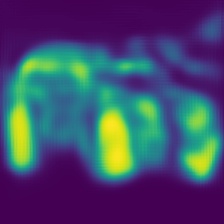}\\
    \includegraphics[width=.98\linewidth, height=2.0cm]{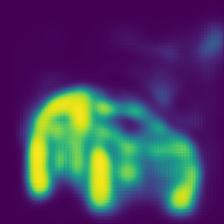} \\
	\vspace{0.15cm}
    \includegraphics[width=.98\linewidth, height=2.0cm]{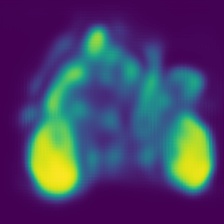}\\
    \includegraphics[width=.98\linewidth, height=2.0cm]{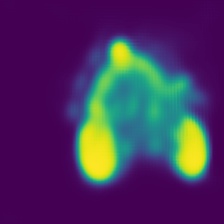}\\
	\vspace{0.15cm}
    \includegraphics[width=.98\linewidth, height=2.0cm]{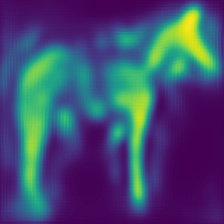}\\
    \includegraphics[width=.98\linewidth, height=2.0cm]{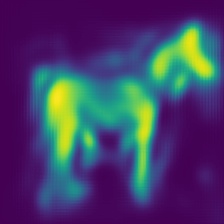}\\
	\vspace{0.15cm}
	\includegraphics[width=.98\linewidth, height=2.0cm]{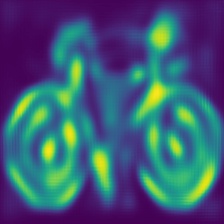}\\
    \includegraphics[width=.98\linewidth, height=2.0cm]{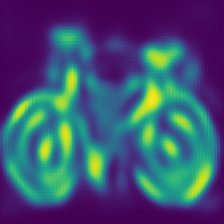}\\
    \vspace{-0.15cm}
	\caption*{\small Ours Prob.}
\end{subfigure}
\hspace{-0.15cm}
\begin{subfigure}[b]{.18\linewidth}
    \centering
    \includegraphics[width=.98\linewidth, height=2.0cm]{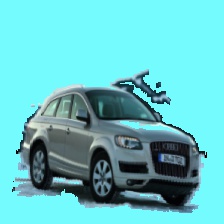} \\
    \includegraphics[width=.98\linewidth, height=2.0cm]{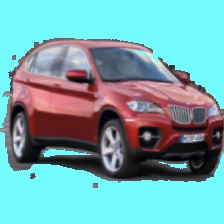} \\
	\vspace{0.15cm}
    \includegraphics[width=.98\linewidth, height=2.0cm]{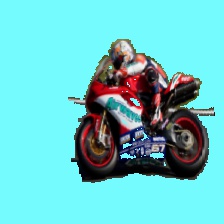} \\
    \includegraphics[width=.98\linewidth, height=2.0cm]{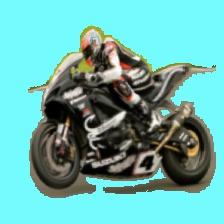} \\
	\vspace{0.15cm}
    \includegraphics[width=.98\linewidth, height=2.0cm]{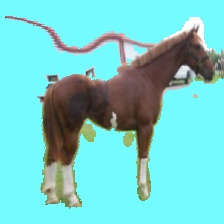} \\
    \includegraphics[width=.98\linewidth, height=2.0cm]{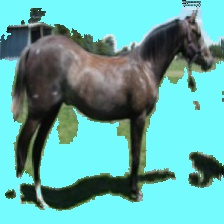} \\
	\vspace{0.15cm}
	\includegraphics[width=.98\linewidth, height=2.0cm]{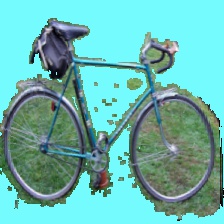} \\
    \includegraphics[width=.98\linewidth, height=2.0cm]{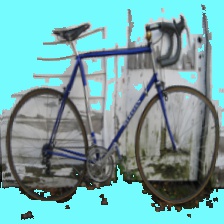} \\
    \vspace{-0.15cm}
    \caption*{\small Ours}
\end{subfigure}
\hspace{-0.15cm}
\begin{subfigure}[b]{.18\linewidth}
    \centering
    \includegraphics[width=.98\linewidth, height=2.0cm]{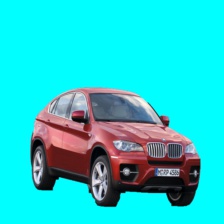} \\
    \includegraphics[width=.98\linewidth, height=2.0cm]{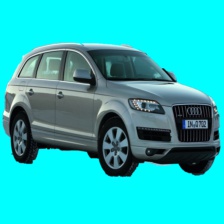} \\
	\vspace{0.15cm}
	\includegraphics[width=.98\linewidth, height=2.0cm]{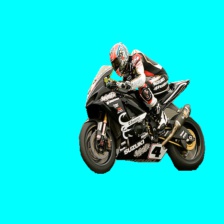} \\
    \includegraphics[width=.98\linewidth, height=2.0cm]{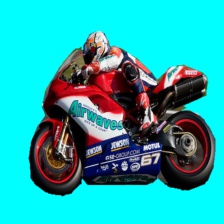} \\
	\vspace{0.15cm}
	\includegraphics[width=.98\linewidth, height=2.0cm]{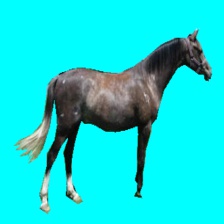} \\
    \includegraphics[width=.98\linewidth, height=2.0cm]{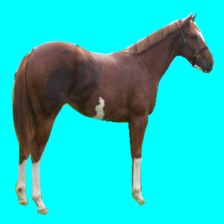} \\
	\vspace{0.15cm}
	\includegraphics[width=.98\linewidth, height=2.0cm]{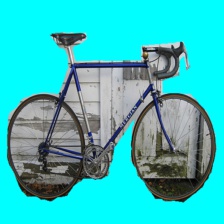} \\
    \includegraphics[width=.98\linewidth, height=2.0cm]{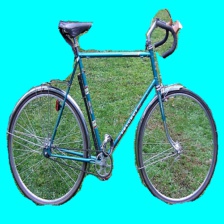} \\
    \vspace{-0.15cm}
	\caption*{\small Target Seg.} 
\end{subfigure}
\hspace{-0.15cm}
\begin{subfigure}[b]{.18\linewidth}
    \centering
    \includegraphics[width=.98\linewidth, height=2.0cm]{supplement/tss_seg/0080_image_2.jpg} \\
    \includegraphics[width=.98\linewidth, height=2.0cm]{supplement/tss_seg/0080_image_1.jpg} \\
	\vspace{0.15cm}
	\includegraphics[width=.98\linewidth, height=2.0cm]{supplement/tss_seg/0356_image_2.jpg} \\
    \includegraphics[width=.98\linewidth, height=2.0cm]{supplement/tss_seg/0356_image_1.jpg} \\
	\vspace{0.15cm}
	\includegraphics[width=.98\linewidth, height=2.0cm]{supplement/tss_seg/0219_image_2.jpg} \\
    \includegraphics[width=.98\linewidth, height=2.0cm]{supplement/tss_seg/0219_image_1.jpg} \\
	\vspace{0.15cm}
	 \includegraphics[width=.98\linewidth, height=2.0cm]{supplement/tss_seg/0344_image_2.jpg} \\
    \includegraphics[width=.98\linewidth, height=2.0cm]{supplement/tss_seg/0344_image_1.jpg} \\
    \vspace{-0.15cm}
	\caption*{\small Target}
\end{subfigure}
\vspace{-0.15cm}
\caption[]{ \small 
Qualitative examples on the Taniai dataset \cite{Taniai2016JointImagesb},
which shows a source image, the probability map of our segmentation network (Ours Prob.), the transformation of our semantic flow network with segmentation (Ours), the ground truth segmentation of the target (Target Seg.) and the target image.
}
\label{fig:resultsTaniai1}
\end{figure*}

%
\begin{figure*}[!t]
\centering
\begin{subfigure}[b]{.18\linewidth}
    \centering
    \includegraphics[width=.98\linewidth, height=2.0cm]{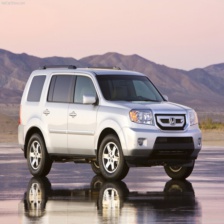} \\
    \includegraphics[width=.98\linewidth, height=2.0cm]{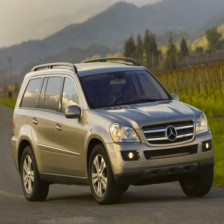} \\
	\vspace{0.15cm}
    \includegraphics[width=.98\linewidth, height=2.0cm]{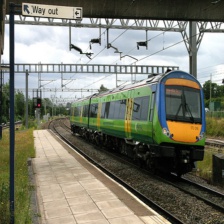} \\
    \includegraphics[width=.98\linewidth, height=2.0cm]{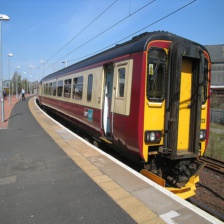} \\
	\vspace{0.15cm}
    \includegraphics[width=.98\linewidth, height=2.0cm]{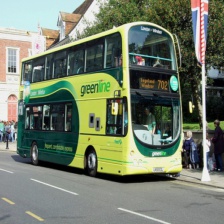} \\
    \includegraphics[width=.98\linewidth, height=2.0cm]{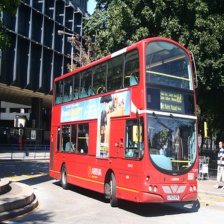} \\
	\vspace{0.15cm}
	\includegraphics[width=.98\linewidth, height=2.0cm]{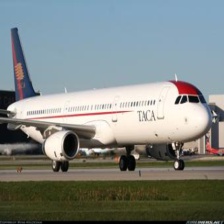} \\
    \includegraphics[width=.98\linewidth, height=2.0cm]{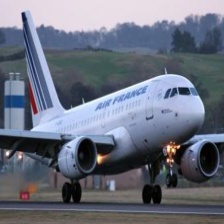} \\
    \vspace{-0.15cm}
	\caption*{\small Source} 
\end{subfigure}
\hspace{-0.15cm}
\begin{subfigure}[b]{.18\linewidth}
    \centering
    \includegraphics[width=.98\linewidth, height=2.0cm]{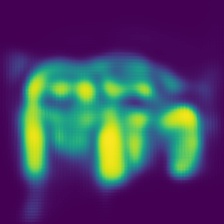}\\
    \includegraphics[width=.98\linewidth, height=2.0cm]{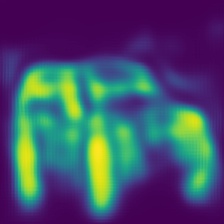}\\
	\vspace{0.15cm}
    \includegraphics[width=.98\linewidth, height=2.0cm]{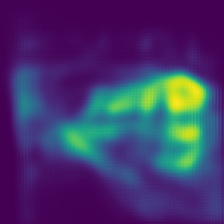}\\
    \includegraphics[width=.98\linewidth, height=2.0cm]{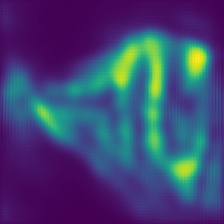}\\
	\vspace{0.15cm}
    \includegraphics[width=.98\linewidth, height=2.0cm]{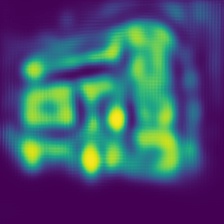}\\
    \includegraphics[width=.98\linewidth, height=2.0cm]{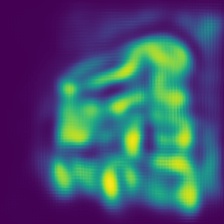}\\
	\vspace{0.15cm}
	\includegraphics[width=.98\linewidth, height=2.0cm]{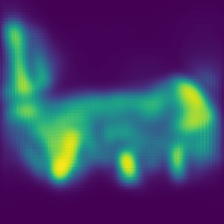}\\
    \includegraphics[width=.98\linewidth, height=2.0cm]{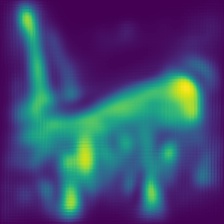}\\
    \vspace{-0.15cm}
	\caption*{\small Ours Prob.}
\end{subfigure}
\hspace{-0.15cm}
\begin{subfigure}[b]{.18\linewidth}
    \centering
    \includegraphics[width=.98\linewidth, height=2.0cm]{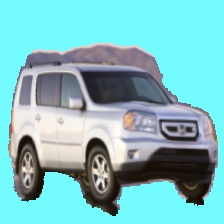} \\
    \includegraphics[width=.98\linewidth, height=2.0cm]{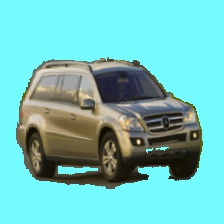} \\
	\vspace{0.15cm}
    \includegraphics[width=.98\linewidth, height=2.0cm]{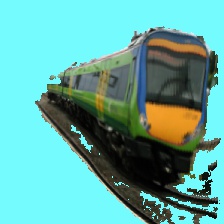} \\
    \includegraphics[width=.98\linewidth, height=2.0cm]{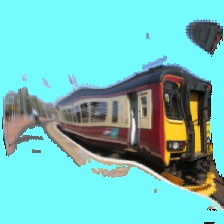} \\
	\vspace{0.15cm}
    \includegraphics[width=.98\linewidth, height=2.0cm]{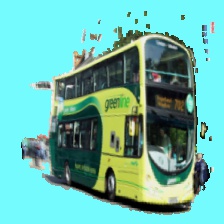} \\
    \includegraphics[width=.98\linewidth, height=2.0cm]{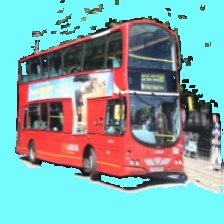} \\
	\vspace{0.15cm}
	\includegraphics[width=.98\linewidth, height=2.0cm]{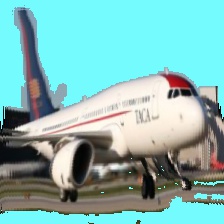} \\
    \includegraphics[width=.98\linewidth, height=2.0cm]{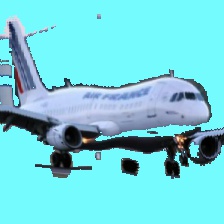} \\
    \vspace{-0.15cm}
    \caption*{\small Ours}
\end{subfigure}
\hspace{-0.15cm}
\begin{subfigure}[b]{.18\linewidth}
    \centering
    \includegraphics[width=.98\linewidth, height=2.0cm]{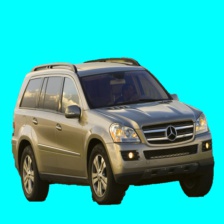} \\
    \includegraphics[width=.98\linewidth, height=2.0cm]{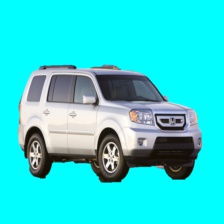} \\
	\vspace{0.15cm}
	\includegraphics[width=.98\linewidth, height=2.0cm]{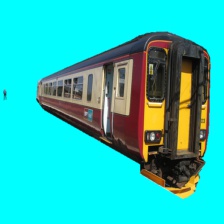} \\
    \includegraphics[width=.98\linewidth, height=2.0cm]{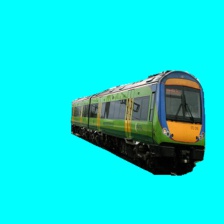} \\
	\vspace{0.15cm}
	\includegraphics[width=.98\linewidth, height=2.0cm]{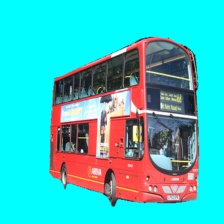} \\
    \includegraphics[width=.98\linewidth, height=2.0cm]{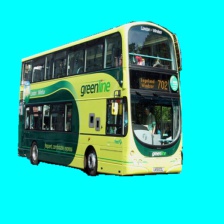} \\
	\vspace{0.15cm}
	\includegraphics[width=.98\linewidth, height=2.0cm]{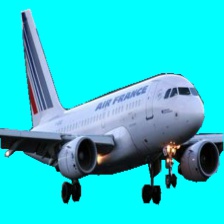} \\
    \includegraphics[width=.98\linewidth, height=2.0cm]{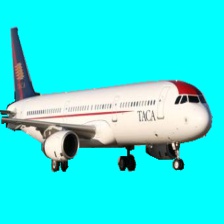} \\
    \vspace{-0.15cm}
	\caption*{\small Target Seg.} 
\end{subfigure}
\hspace{-0.15cm}
\begin{subfigure}[b]{.18\linewidth}
    \centering
    \includegraphics[width=.98\linewidth, height=2.0cm]{supplement/tss_seg/0037_image_2.jpg} \\
    \includegraphics[width=.98\linewidth, height=2.0cm]{supplement/tss_seg/0037_image_1.jpg} \\
	\vspace{0.15cm}
	\includegraphics[width=.98\linewidth, height=2.0cm]{supplement/tss_seg/0366_image_2.jpg} \\
    \includegraphics[width=.98\linewidth, height=2.0cm]{supplement/tss_seg/0366_image_1.jpg} \\
	\vspace{0.15cm}
	\includegraphics[width=.98\linewidth, height=2.0cm]{supplement/tss_seg/0277_image_2.jpg} \\
    \includegraphics[width=.98\linewidth, height=2.0cm]{supplement/tss_seg/0277_image_1.jpg} \\
	\vspace{0.15cm}
	 \includegraphics[width=.98\linewidth, height=2.0cm]{supplement/tss_seg/0229_image_2.jpg} \\
    \includegraphics[width=.98\linewidth, height=2.0cm]{supplement/tss_seg/0229_image_1.jpg} \\
    \vspace{-0.15cm}
	\caption*{\small Target}
\end{subfigure}
\vspace{-0.15cm}
\caption[]{ \small 
Qualitative examples on the Taniai dataset \cite{Taniai2016JointImagesb},
which shows a source image, the probability map of our segmentation network (Ours Prob.), the transformation of our semantic flow network with segmentation (Ours), the ground truth segmentation of the target (Target Seg.) and the target image.
}
\label{fig:resultsTaniai2}
\end{figure*}

\end{document}